\documentclass[preprint,12pt]{elsarticle}

\usepackage{amsmath,amsfonts}
\usepackage{amssymb}

\usepackage[ruled,vlined]{algorithm2e}
\usepackage{float}
\usepackage{placeins}

\usepackage{multirow}
\usepackage{makecell}
\usepackage{array}

\usepackage{subcaption}

\usepackage{longtable}
\usepackage{adjustbox}

\usepackage{enumitem}

\usepackage[colorlinks=true,citecolor=blue,linkcolor=blue]{hyperref}

\begin{document}

\begin{frontmatter}

\title{An Attention-Guided Global and Local Fusion Framework for Lesion-Focused Image Classification}

\author[ju]{Mst Shafia Tasnim}

\author[ju]{Md Samaun Elahee}

\author[uftb]{Tanjim Taharat Aurpa\corref{cor1}}
\ead{aurpa0001@uftb.ac.bd}

\author[ju]{Md Musfique Anwar}

\cortext[cor1]{Corresponding author}

\address[ju]{Department of Computer Science and Engineering,
	Jahangirnagar University, Savar, Bangladesh}

\address[uftb]{Department of Data Science and Engineering,
	University of Frontier Technology, Bangladesh (UFTB),
	Gazipur, Bangladesh}

\begin{abstract}

Lesion-focused image classification presents a core analytical challenge, as discriminative signals are often sparse, spatially dispersed, and easily obscured by background noise, while conventional convolutional neural networks (CNNs) process entire images uniformly and may dilute signal relevance. This study hypothesizes that adaptive fusion of global contextual information and lesion-focused local information can improve classification performance compared with using either representation independently. 
We propose a three-branch, attention-guided deep learning framework built on Densely Connected Convolutional Network-121 (DenseNet-121) to improve feature attribution, interpretability, and classification reliability. The architecture consists of a global branch that learns representations from full images, followed by Gradient-weighted Class Activation Mapping (Grad-CAM) to generate attention maps that highlight prediction-relevant regions and produce masked inputs, and a local branch enhanced with a Convolutional Block Attention Module (CBAM) to extract refined spatial and channel-wise features from these focused regions. An adaptive fusion branch integrates global and local representations by learning instance-specific weights, allowing dynamic prioritization between contextual and localized information. 
The framework is evaluated on a synthetic Spot Pattern Dataset (SSPD) and three benchmark datasets, including skin lesion, guava leaf, and grape leaf image datasets, where the fusion branch outperformed the individual global and local branches,
reaching 97.75\% accuracy on the skin lesion dataset and 99.64\% on the guava leaf dataset.

The results highlight the value of attention-guided architectures in healthcare analytics by improving model transparency, strengthening feature relevance, and supporting more reliable data-driven decision-making in medical image analysis.

\end{abstract}





\end{frontmatter}

\section{Introduction}\label{sec1}

Many visual classification or identification tasks involve the presence of small and scattered regions of interest where discriminative information is limited to localized patterns, rather than the global image structure. Such lesion- or spot-focused patterns are common across diverse application domains, including skin lesion image classification \cite{Kim2023,Asif2025,Thurnhofer-Hemsi2021,Wang2022}, plant leaf disease classification \cite{Kunduracioglu2024a,Praveen2023a,Prasad2024}, fruit disease imagery for classification\cite{Shihab2025}, etc. Deep convolutional neural networks have demonstrated strong performance in such classification tasks due to their ability to automatically learn hierarchical feature representations \cite{Amin2022a,Hoang2022a,A2024,Nigar2022}. However, learning representations that jointly capture broad image context and subtle, spatially scattered discriminative cues for accurate classification remains challenging.

Models trained on full images may face challenges in lesion-centric classification tasks, as discriminative lesion cues are often small and localized and may not be sufficiently captured in global image representations. In real-world datasets where lesions exhibit significant variation in size, shape, and spatial distribution, relying solely on global image representations may therefore be insufficient for accurately capturing fine-grained lesion-specific patterns.

To address this, many studies have adopted region-of-interest (ROI)–focused strategies, where ROIs are first segmented or localized and then used for classification \cite{Ahammed2022,Alshahrani2024}. ROI extraction has been performed using diverse techniques, including thresholding-based segmentation approaches, EW-FCM (Entropy-Weighted First-Order Cumulative Moment) \cite{Hoang2022a}, deep learning–based instance segmentation such as Mask Region-based Convolutional Neural Network (MRCNN) \cite{Akram2023}, U-Net-based segmentation followed by watershed-based ROI extraction \cite{mondal2025artificial}, and post hoc interpretability tools like Grad-CAM and saliency maps \cite{Ashwath2023a}. While these approaches enhance lesion focus, they typically discard surrounding contextual information, which can be crucial for reliable prediction. Moreover, their performance is often sensitive to segmentation quality, as errors in ROI extraction may propagate to subsequent classification stages. These limitations indicate the need for models that can jointly leverage global context and localized lesion representations.

In this work, we propose a global–local fusion framework that integrates attention-guided localization with adaptive feature fusion to address these challenges. The global branch extracts feature representations from the full image, capturing contextual and structural information. Then, Grad-CAM is used to obtain heatmaps that identify prediction-relevant regions, which are used to generate attention-guided masked images. The local branch subsequently extracts features from the lesion-centric images, emphasizing discriminative lesion patterns. Finally, an adaptive weighting fusion module dynamically balances global and local feature contributions during prediction, allowing the model to prioritize global context or localized cues based on the characteristics of each input image. This design allows global and local representations to contribute differently across images with varying lesion visibility and scale. Our main contributions are as follows:

\begin{itemize}[label=\textbullet]

    \item We address lesion-centric visual classification involving small, sparsely distributed, and spatially scattered discriminative regions by proposing a global–local fusion framework that jointly models full-image contextual information and lesion-focused representations.
    
    \item Through an adaptive weighting strategy, the proposed framework dynamically balances global contextual features and localized lesion cues, allowing the model to prioritize the most relevant information according to the characteristics of each input image.
    
    \item The proposed framework is evaluated on one synthetic dataset and three real-world datasets from medical and agricultural domains. This experimental setup allows a broader evaluation of the model across diverse lesion appearances, spatial distributions, and background conditions. In addition, the impact of background complexity is analyzed to assess the framework’s sensitivity to background variations, providing insights into its limitations and potential directions for further improvement.

\end{itemize}

The rest of this paper is organized as follows. Section 2 reviews related work under three categories: global–local feature fusion, lesion-focused classification, and attention/heatmap-guided learning. Section 3 describes the proposed methodology, followed by experimental results in Section 4. Section 5 provides the discussion, and Section 6 concludes the paper.

\section{Related Work}\label{sec2}

\subsection*{Global-local feature fusion}

In \cite{Ashwath2023a}, Ashwath et al. proposed a three-tier self-interpretable architecture consisting of a global branch, an attention branch, and a local branch for medical image classification. The global branch is trained on original images. Attention maps are generated to create masked images for local branch input. Finally, the pooled features from global and local branches are concatenated and fed to a fusion network for final prediction. While this setup improves interpretability and performance, the local branch in the synthetic blob dataset is initialized with the global branch’s weights.  In our experiments, we observed that independently training the local branch resulted in better fusion performance, suggesting that the local pathway learns more complementary region-focused representations when it is not initialized from the trained global branch weights. Moreover, we incorporated an adaptive attention-based fusion mechanism that learns sample-wise importance weights instead of relying on simple feature concatenation. In \cite{Guan2018a}, Guan et al. proposed AG-CNN, which also consists of three branches. The global branch captures information from the entire image, while the local branch focuses on the disease-relevant regions highlighted by the attention mechanism. Global and local features are then concatenated. The effectiveness of AG-CNN is demonstrated for thorax disease classification. However, the framework selects only a single dominant connected region, limiting its ability to model disconnected or spatially dispersed disease patterns common in lesion-focused analysis.

In \cite{prasad2023explainable}, the authors proposed a binary lung disease classification framework in which XAI was used to identify ROI regions from chest X-ray images. Features were extracted from both the full image and the ROI-based local regions. In the final stage, the global and local features were concatenated, with equal weight assigned during fusion for final classification. The local and fusion models achieved improved performance, reporting an accuracy of 99.6\% with fewer training epochs. However, assigning equal contribution to global and local features may limit fusion flexibility, whereas our framework uses adaptive fusion to learn the relative importance of each branch for each input sample. 

Wang et al. \cite{Wang2023a} proposed DHUnet, a dual-branch hierarchical global–local fusion network for whole-slide image segmentation that integrates Swin Transformer and ConvNeXt modules and fuses global coarse and local fine-grained features through multi-scale fusion modules. Lian et al.\cite{Lian2023a} proposed a three-branch local–global feature fusion network. The method extracts local texture and global semantic information from the same image and fuses them for final classification. Experiments on the MSTAR dataset report average recognition accuracies of 99.26\% under standard operating conditions. In contrast, our framework separates the two representations explicitly by learning global context from the full image and local lesion-specific features from the ROI-guided masked image. 

In \cite{Li2025a}, the global branch employs ResNet50 with Multi-Head self-attention to capture global context, while the local branch focuses on fine-grained features from the segmented tumor region, and an attention-enhanced fusion module is used to filter and integrate important features. Although the method achieved strong classification performance, its local branch relies on pre-existing expert-annotated tumor segmentation masks to extract local tumor regions, which may limit its applicability when such masks are unavailable. In contrast, our framework derives prediction-relevant local inputs from the trained global branch using Grad-CAM, enabling local feature learning without requiring separate pixel-level mask annotations.

Liang et al. \cite{Liang2024a} proposed GLSNet, a dual-branch framework that combines global context from full images with local details extracted from cropped patches using a transformer-based fusion mechanism. The method is designed for ultra-high-resolution image segmentation and achieved a 0.8\% improvement in segmentation accuracy over the baseline model. 
In \cite{Xing2020a}, the authors proposed AGDN, where attention maps generated by the first branch are used to zoom in on lesion regions in the input images of the second branch. In our approach, we have explicitly removed background information and trained the local branch using lesion-only masked images, thereby reducing background interference and enabling more focused lesion-centric feature learning. AGDN further incorporates third-order long-range feature aggregation modules to capture global contextual information and employs a deformation-based attention consistency loss to refine attention maps and encourage cross-branch consistency. The global feature embeddings from both branches are fused for final classification, achieving an overall accuracy of 91.29\% on two public wireless capsule endoscopy (WCE) datasets. In our model, we adopt an adaptive fusion strategy to dynamically integrate global and local representations.

\subsection*{Lesion-focused classification}

In lesion-centric classification, the target classes are primarily defined by lesion characteristics rather than a global image appearance. To guide the model’s attention towards the lesion region, many studies include removing the background from the image and train the model using the segmented lesion area only, while some approaches extract features from full images without explicit lesion segmentation.

Authors in \cite{Ogudo2022a} introduced a classification framework where the affected region is segmented using multi-level thresholding. Then, features are extracted from the segmented lesion area using an optimized stacked sparse autoencoder. The model achieved 94.7\% accuracy on a 7-class skin lesion classification task. However, the method uses threshold-based segmentation, and features are extracted from the segmented lesion region; therefore, its classification accuracy is sensitive to the correctness of the segmentation result. In contrast, our framework uses both the full image and the lesion-focused ROI, preserving global background context while emphasizing the affected region.

In \cite{Deepa2025a}, the authors proposed an Adaptive Layer-based Visual Transformer with UNet (ALVTransUNet), where the segmented images are fed into Dilated DenseNet with a Multi-Head Attention mechanism (DD-MHA) for classification. Since the classification stage is based on segmentation outputs, the final accuracy is sensitive to the correctness of lesion localization.

In \cite{Hoang2022a}, the authors used entropy-based weighting and first-order cumulative moments to generate a segmented lesion image; the resulting image highlights and delineates the lesion boundaries while retaining the background, and is fed into a wide-ShuffleNet model for classification. Aldhyani et al. \cite{Aldhyani2022a} proposed a lightweight CNN with dynamic kernel sizes for multi-class skin lesion classification, achieving an overall accuracy of 97.85\% across seven classes on the HAM10000 dataset. Elfatimi et al. \cite{Elfatimi2022a} investigated lightweight MobileNet architectures for bean leaf disease classification. Authors in \cite{Shetty2022a} presented a customized convolutional neural network trained on skin lesion images, reporting a maximum accuracy of 95.18\% across seven lesion categories.

In contrast to the approaches that rely solely on segmented lesion regions or exclusively on full-image representations, our method achieves lesion-focused classification by explicitly combining global contextual information from the entire image with localized lesion-specific features. Rather than discarding background information, the proposed framework prioritizes lesion regions while retaining complementary contextual cues, enabling robust classification even in scenarios where lesion appearance is ambiguous or spatially distributed.

\subsection*{Attention/Heatmap Guided Learning}

Mohammed et al. \cite{Mohammed2023b} employed a Grad-CAM–guided preprocessing strategy where heatmaps are used to crop the region of interest (ROI) from mammograms for classification.
Authors in \cite{Haut2019b} proposed an attention-driven approach for hyperspectral image classification, where a mask is computed and applied to the extracted feature maps to highlight the most informative features.

Authors in \cite{Liao2019} proposed EAMNet, a clinically interpretable ConvNet architecture that aggregates multi-scale features using Multi-Layers Average Pooling (M-LAP) and employs Evidence Activation Mapping to highlight diagnostically relevant regions. The method achieved an AUC of 0.88. 
In  \cite{Eslami2021}, the authors proposed A+MCNN, an attention-based multi-scale CNN that encodes contextual information using multi-scale input tiles and integrates features through a mid-fusion strategy with an attention module for improved pavement image classification. In \cite{abugabah2025deep}, the author proposed a deep learning framework for automated breast cancer diagnosis using color normalization and attention-guided segmentation, followed by feature extraction and classification. Their model achieved high diagnostic performance on the WBCD and BreakHis datasets.

The authors in \cite{Yang2020a} employ region-level supervision to localize the ROI and guide the attention during classification, activating features in diagnostically relevant regions while reducing the influence of irrelevant background.

Thus, in some of the above studies, attention mechanisms are applied internally
at the feature level and are not explicitly designed to provide region level
visual localization. Several of them employ attention mainly within intermediate feature representations. In contrast, our framework leverages Grad-CAM
to generate heatmaps and corresponding masked images, enhancing visual
interpretability while guiding lesion-focused learning. Furthermore, CBAM is integrated into the local branch to enhance spatial and channel-wise feature refinement. Additionally, in our model, prediction-relevant regions are derived from the last convolutional layer of the global branch without requiring region-level supervision, which is often unavailable due to the need for expert annotation.

\section{Methodology}\label{sec3}

\begin{figure}[ht]
	\centering
	\includegraphics[width=0.9\textwidth]{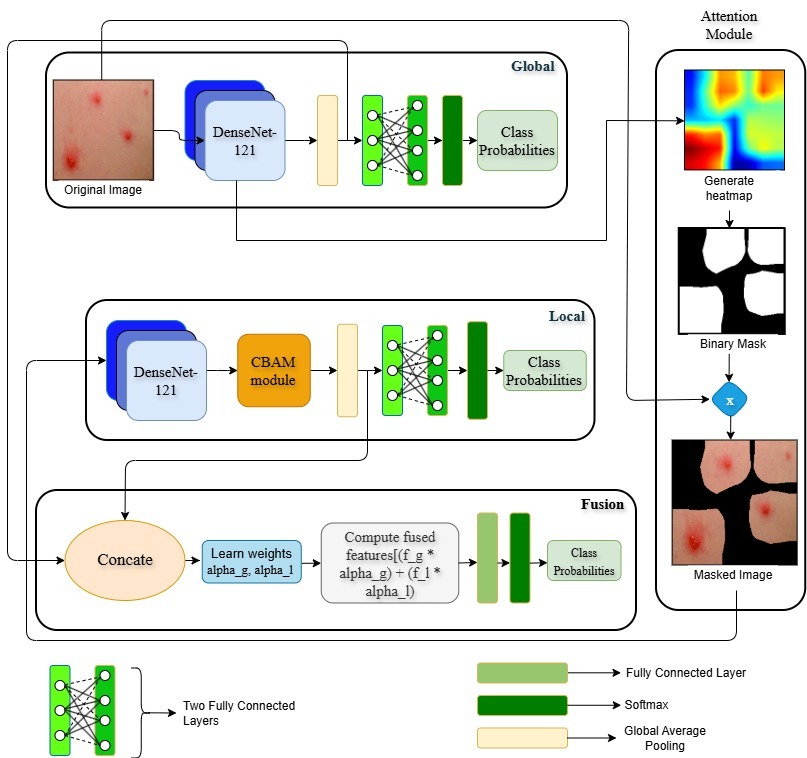} 
	\caption{System architecture of the proposed framework}
	\label{fig:overall_architecture}
\end{figure}

Lesion-focused image classification is challenging because discriminative evidence is often confined to small, spatially scattered regions, while background and global structures may dominate the image. To address this, we propose a global–local fusion framework that jointly models full-image contextual information and lesion-centric details within a unified architecture.
\newline
The framework is designed to first understand the image globally, then progressively refine its focus toward discriminative regions, and finally combine both sources of information adaptively for robust classification. Instead of relying solely on full-image representations or exclusively on segmented regions, the proposed model dynamically balances global context and localized features based on the characteristics of each input image. An overview of the proposed architecture is illustrated in Fig.~\ref{fig:overall_architecture}. The framework comprises three branches: a global branch, a local branch, a fusion branch, along with an attention module. The global branch captures high-level feature representations from the full image and produces initial predictions. The feature map from the final convolutional layer and class-specific gradients are subsequently used in the attention module, where a post-hoc visualization technique, Gradient-weighted Class Activation Mapping (Grad-CAM), generates a heatmap that highlights salient regions driving the classification decision. The heatmap is then thresholded to generate a binary mask, which is applied to the original image to remove irrelevant background regions and produce focused, masked images.
\newline
The local branch is trained on these masked images to capture fine-grained, class-specific features within the most informative regions. To further refine feature representation, it incorporates a Convolutional Block Attention Module (CBAM) that emphasizes informative spatial and channel-wise patterns within the masked regions. Finally, the fusion branch integrates the global and local feature representations using an adaptive weighting strategy that learns the relative contribution of each branch. This fusion mechanism enables the model to balance global contextual information and localized lesion-specific features on a per-sample basis and produce the final classification. 
\newline
Depending on lesion size, appearance, and spatial distribution, some images benefit more from global contextual cues while others rely more heavily on localized details. Therefore, the adaptive fusion mechanism flexibly balances global and local representations under varying lesion visibility, leading to more stable and reliable predictions.

\vspace{1.5em}

\noindent\textbf{Generalized Notation.} The key components of the framework are summarized as follows:

\begin{itemize}
	\item $I$: Original input image.
	\item $I_{\mathrm{masked}}$: Masked image generated using binary masking Grad-CAM.
	\item $\mathbf{f}_g$: Global feature vector extracted from $I$ by the global branch.
	\item $\mathbf{f}_l$: Local feature vector extracted from $I_{\mathrm{masked}}$ by the Convolutional Block Attention Module (CBAM)-enhanced local branch.
	\item $[\alpha_g, \alpha_l]$: Adaptive fusion weights learned from global and local features, with $\alpha_g + \alpha_l = 1$.
	\item $\mathbf{f}_{\mathrm{fused}} = \alpha_g \mathbf{f}_g + \alpha_l \mathbf{f}_l$: Final fused feature representation.
	\item $\hat{\mathbf{y}}$: Final predicted class probability vector.
	
\end{itemize}

\subsection{Global Branch}

The input to the global branch consists of full images $I$, which are resized to $224 \times 224$ and normalized. We employ DenseNet-121, pretrained on ImageNet, as the backbone network and fine-tune it by unfreezing the last 50 layers. Let $f_{\theta_g}(\cdot)$ denote the DenseNet-121-based global feature extractor with learnable parameters $\theta_g$. The feature extraction process is defined in Eq.~\eqref{eq1}:
\begin{equation}
	\label{eq1}
	\mathbf{f}_g = f_{\theta_g}(I),
\end{equation}
where $\mathbf{f}_g \in \mathbb{R}^{1024}$ represents the feature vector obtained after the Global Average Pooling (GAP) layer.

The extracted global feature vector is further transformed through two fully connected layers with 256 and 128 units, respectively, for classification, as shown in Eq.~\eqref{eq2}.

\begin{equation}
	\label{eq2}
	\begin{aligned}
		h_1 &= \mathrm{Dropout}\Big(\mathrm{BN}\big(\delta(W_1 \mathbf{f}_g + b_1)\big)\Big), \\
		h_2 &= \mathrm{Dropout}\Big(\mathrm{BN}\big(\delta(W_2 h_1 + b_2)\big)\Big), \\
		\hat{\mathbf{y}}_g &= \mathrm{softmax}(W_3 h_2 + b_3),
	\end{aligned}
\end{equation}
where \(h_1\) and \(h_2\) denote hidden representations from the first and second fully connected layers, respectively; \(W_1, W_2, W_3\) and \(b_1, b_2, b_3\) are learnable weights and biases; \(\delta(\cdot)\) denotes the ReLU activation function; BN denotes Batch Normalization; and \(\hat{\mathbf{y}}_g\) represents the predicted class probability vector of the global branch.

The global model is trained using categorical cross-entropy loss, as defined in Eq.~\eqref{eq3}.
\begin{equation}
	\label{eq3}
	\mathcal{L}_{\mathrm{global}} = -\sum_{k=1}^{C} y^{(k)} \log \hat{\mathbf{y}}_{g}^{(k)},
\end{equation}
where \(C\) is the number of target classes, \(y^{(k)}\) is the one-hot encoded ground-truth label for class \(k\), and \(\hat{\mathbf{y}}_{g}^{(k)}\) is the predicted probability of the global branch for class \(k\).

After training the global model, its learned representations are used to generate Grad-CAM heatmaps and to provide global features for the fusion module.

\subsection{Attention Module}

To localize the most discriminative regions guiding the model’s prediction, we employ a post hoc, class-discriminative localization module based on Gradient-weighted Class Activation Mapping (Grad-CAM). Let \( I \in \mathbb{R}^{224 \times 224 \times 3} \) denote an input image resized and normalized before being forwarded through the trained global branch \( f_{\theta_g} \). The global branch produces class probabilities, and 
Grad-CAM is computed with respect to the predicted class \(c=\arg\max_k \hat{\mathbf{y}}_g^{(k)}\).

Grad-CAM generates a localization heatmap by computing the gradient of the predicted class probability with respect to the feature maps of the last convolutional layer of the global branch. These gradients capture the contribution of each spatial location to the model’s decision. The gradients are globally averaged to obtain channel-wise importance weights, which are then used to compute a weighted combination of the convolutional feature maps followed by a ReLU activation. The resulting heatmap \( H^c \) is resized to \(224 \times 224\) and normalized to the range \([0,1]\).

To isolate lesion-relevant regions, a binary mask \(M \in \{0,1\}^{224 \times 224}\) is constructed using a threshold value of \(\tau = 0.5\), as defined in Eq.~\eqref{eq4}.
\begin{equation}
	\label{eq4}
	M(i,j) =
	\begin{cases}
		1, & \text{if } H^c(i,j) > \tau, \\
		0, & \text{otherwise}
	\end{cases}
\end{equation}
where \(H^c(i,j)\) denotes the normalized Grad-CAM activation value at spatial location \((i,j)\) for class \(c\), and \(\tau\) denotes the masking threshold.

The binary mask is then applied to the input image through pixel-wise background suppression. The masked image \(I_{\text{masked}}\) is obtained using Eq.~\eqref{eq5}.
\begin{equation}
	\label{eq5}
	I_{\mathrm{masked}}(i,j,r) =
	\begin{cases}
		I(i,j,r), & \text{if } M(i,j)=1, \\
		0, & \text{otherwise}
	\end{cases}
\end{equation}
where \(i\) and \(j\) denote spatial coordinates, and \(r\) denotes the image channel index.

This operation removes background regions while preserving class-discriminative visual cues identified by the global branch. The resulting masked images are subsequently used to train the local branch, enabling it to focus on fine-grained, lesion-centric features and complement the global contextual representation.

\subsection{Local Branch with CBAM}

The local branch shares the same DenseNet-121 backbone architecture as the global branch and is trained on the masked images generated by the Grad-CAM-based attention module. 
To enhance feature representation within the salient masked regions, a Convolutional Block Attention Module (CBAM) is inserted after the final convolutional feature extraction stage of DenseNet-121 and before the Global Average Pooling layer, allowing the network to focus on discriminative patterns relevant to classification.

Let $F \in \mathbb{R}^{H \times W \times D}$ denote an intermediate feature map extracted from the DenseNet-121 backbone, where $H$, $W$, and $D$ represent the height, width, and number of channels, respectively. CBAM sequentially applies channel attention and spatial attention mechanisms, as described in Eqs.~\eqref{eq6}--\eqref{eq9}.

\textbf{Channel Attention:}
\begin{equation}
	\label{eq6}
	M_c(F) = \sigma\big(\mathrm{MLP}(\mathrm{AvgPool}(F)) + \mathrm{MLP}(\mathrm{MaxPool}(F))\big),
\end{equation}
\begin{equation}
	\label{eq7}
	F' = M_c(F) \odot F,
\end{equation}

where $M_c(F)$ denotes the channel attention map, $\sigma(\cdot)$ denotes the sigmoid activation function, MLP represents a shared multi-layer perceptron, $\mathrm{AvgPool}(\cdot)$ and $\mathrm{MaxPool}(\cdot)$ denote global average pooling and global max pooling, respectively, and $\odot$ denotes element-wise multiplication. The refined feature map $F'$ is obtained by reweighting the input feature map $F$ using the channel attention map.

\textbf{Spatial Attention:}
\begin{equation}
	\label{eq8}
	M_s(F') = \sigma\big(\mathrm{Conv}_{7 \times 7}([\mathrm{Avg}(F'); \mathrm{Max}(F')])\big),
\end{equation}
\begin{equation}
	\label{eq9}
	F'' = M_s(F') \odot F',
\end{equation}
where $M_s(F')$ denotes the spatial attention map, and average and max pooling are computed along the channel dimension and concatenated before the $7 \times 7$ convolution. The refined feature map $F''$ is obtained by applying the spatial attention map to $F'$.

The refined feature map $F''$ is subsequently aggregated using Global Average Pooling (GAP) and passed through fully connected layers followed by a softmax classifier. The resulting output \(\hat{\mathbf{y}}_l\) denotes the predicted class probability vector of the local branch. The local branch is initialized using ImageNet-pretrained weights, without reusing weights from the global branch. The local model is trained for half the number of epochs used for the global model, and its extracted local features are subsequently used in the fusion module.

\subsection{Adaptive Weighted Fusion Branch}

To integrate global contextual features and localized lesion-specific features, an adaptive weighted fusion strategy is employed. Let $\mathbf{f}_g, \mathbf{f}_l \in \mathbb{R}^{d}$ denote the global and local feature vectors extracted from the global and local branches, respectively, where $d$ represents the common feature dimension.
These feature vectors are concatenated and passed through a lightweight gating network composed of fully connected layers with 256 and 128 units, followed by a two-unit softmax layer that learns sample-specific importance weights for the global and local branches. During fusion training, the global and local feature extractors remained partially trainable, with only the final 50 layers unfrozen.

A softmax operation is applied to generate normalized fusion weights $\alpha_g$ and $\alpha_l$ by feeding the concatenated feature vector $[\mathbf{f}_g; \mathbf{f}_l]$ into a lightweight gating network $\phi(\cdot)$, where $\phi(\cdot)$ denotes the learnable gating function. This operation is expressed in Eq.~\eqref{eq10}.
\begin{equation}
	\label{eq10}
	[\alpha_g,\alpha_l] = \mathrm{softmax}\big(\phi([\mathbf{f}_g; \mathbf{f}_l])\big),
\end{equation}
where $\alpha_g,\alpha_l \in [0,1]$ and $\alpha_g + \alpha_l = 1$. The final fused feature representation is computed as shown in Eq.~\eqref{eq11}.
\begin{equation}
	\label{eq11}
	\mathbf{f}_{\mathrm{fused}} = \alpha_g \cdot \mathbf{f}_g + \alpha_l \cdot \mathbf{f}_l.
\end{equation}
This adaptive weighting mechanism allows the model to dynamically balance global contextual information and localized discriminative features on a per-sample basis. The fused feature vector is then passed through a 256-unit fully connected layer, followed by batch normalization, dropout, and a softmax classifier to obtain the final predicted class probability vector $\hat{\mathbf{y}}$.

\subsection{Training Strategy}

We employed a three-stage training strategy.

\vspace{0.5em}

\textbf{Stage 1:} Algorithm~\ref{alg:global_mask} outlines the global model training and masked image generation process. The full image is used to train the global branch for learning discriminative global representations. Grad-CAM is then applied to the predicted class of the global branch to generate the attention map, which is thresholded to obtain a binary mask. Finally, the mask is used to suppress irrelevant background regions and produce the masked image for the next stage.
\vspace{0.5em}

\begin{algorithm}[H]
	\caption{Global model training and masked image generation}
	\label{alg:global_mask}
	\textbf{Input:} Image $I$, label $y$, threshold $\tau$ \\
	\textbf{Output:} Global feature vector $\mathbf{f}_g$, global branch class probabilities $\hat{\mathbf{y}}_g$, masked image $I_{\mathrm{masked}}$ \\[0.4em]
	\textbf{Initialization:} Initialize the global model $f_{\theta_g}$ with ImageNet-pretrained DenseNet-121.\\
	\textbf{Step 1:} Resize and normalize $I$; compute $\mathbf{f}_g=f_{\theta_g}(I)$ (Eq.~\eqref{eq1}); optimize using $\mathcal{L}_{\mathrm{global}}$ (Eq.~\eqref{eq3}).\\
	\textbf{Step 2:} Compute $\hat{\mathbf{y}}_g$ (Eq.~\eqref{eq2}) and predicted class $c$.\\
	\textbf{Step 3:} Compute Grad-CAM heatmap $H^c$ for class $c$.\\
	\textbf{Step 4:} Generate binary mask $M$ using threshold $\tau$ (Eq.~\eqref{eq4}).\\
	\textbf{Step 5:} Obtain $I_{\mathrm{masked}}$ by suppressing pixels where $M=0$ (Eq.~\eqref{eq5}).
\end{algorithm}

\vspace{1.2em}

\textbf{Stage 2:} Algorithm~\ref{alg:local_cbam} presents the local model training process on the Grad-CAM-masked images to capture fine-grained, lesion-focused features. Let \(f_{\theta_l}(\cdot)\) denote the local feature extractor with learnable parameters \(\theta_l\). The masked image is passed through the DenseNet-121 backbone to obtain an intermediate feature map. CBAM is then applied sequentially through channel and spatial attention to refine the feature representation, after which GAP and the classification head are used to produce the local branch class probabilities.

\vspace{1.2em}

\begin{algorithm}[H]
	\caption{Local model training with CBAM}
	\label{alg:local_cbam}
	\textbf{Input:} Masked image $I_{\mathrm{masked}}$, ground-truth label $y$ \\
	\textbf{Output:} Local feature vector $\mathbf{f}_l$, local branch class probabilities $\hat{\mathbf{y}}_l$ \\[0.4em]
	
	\textbf{Initialization:} Initialize the local model $f_{\theta_l}$ with ImageNet-pretrained DenseNet-121 and integrate CBAM.\\
	\textbf{Step 1:} Forward $I_{\mathrm{masked}}$ through the DenseNet-121 backbone to obtain feature map $F$.\\
	\textbf{Step 2:} Compute channel attention map $M_c(F)$ (Eq.~\eqref{eq6}) and obtain refined feature map $F'$ (Eq.~\eqref{eq7}).\\
	\textbf{Step 3:} Compute spatial attention map $M_s(F')$ (Eq.~\eqref{eq8}) and obtain refined feature map $F''$ (Eq.~\eqref{eq9}).\\
	\textbf{Step 4:} Apply GAP to $F''$ to obtain $\mathbf{f}_l$, and pass $\mathbf{f}_l$ through the classification head to compute $\hat{\mathbf{y}}_l$.\\
	\textbf{Step 5:} Optimize the local model using categorical cross-entropy loss between $y$ and $\hat{\mathbf{y}}_l$.
\end{algorithm}

\textbf{Stage 3:} Algorithm~\ref{alg:fusion} presents the adaptive weighted fusion of global and local features for final classification. The feature representations from the global and local branches are first combined through concatenation. Adaptive fusion weights are then computed to determine the relative contribution of each branch, and the fused representation is finally passed through the classification layers to produce the final class probability vector.

\vspace{0.5em}

\begin{algorithm}[H]
	\caption{Adaptive weighted fusion model training}
	\label{alg:fusion}
	
	\textbf{Input:} Full image $I$, masked image $I_{\mathrm{masked}}$, ground-truth label $y$ \\
	\textbf{Output:} Final class probability vector $\hat{\mathbf{y}}$ \\
	
	\textbf{Initialization:} Load the trained global and local feature extractors. \\

	\textbf{Step 1:} Obtain the global and local feature vectors, $\mathbf{f}_g$ and $\mathbf{f}_l$, from the trained global and local feature extractors.\\

	\textbf{Step 2:} Concatenate the global and local feature vectors and compute fusion weights $[\alpha_g,\alpha_l]$ using Eq.~\eqref{eq10}.\\

	\textbf{Step 3:} Compute the fused feature representation $\mathbf{f}_{\mathrm{fused}}$ using Eq.~\eqref{eq11}. \\
	
	\textbf{Step 4:} Pass $\mathbf{f}_{\mathrm{fused}}$ through the classification layers to obtain the final class probability vector $\hat{\mathbf{y}}$. \\
	
	\textbf{Step 5:} Optimize the fusion model using categorical cross-entropy loss between $y$ and $\hat{\mathbf{y}}$. \\
	
\end{algorithm}
\vspace{0.5em}

\section{Results}\label{sec4}

\subsection{Dataset Description}
\begin{figure}[htbp] 
    \centering
    \includegraphics[width=0.9\textwidth]{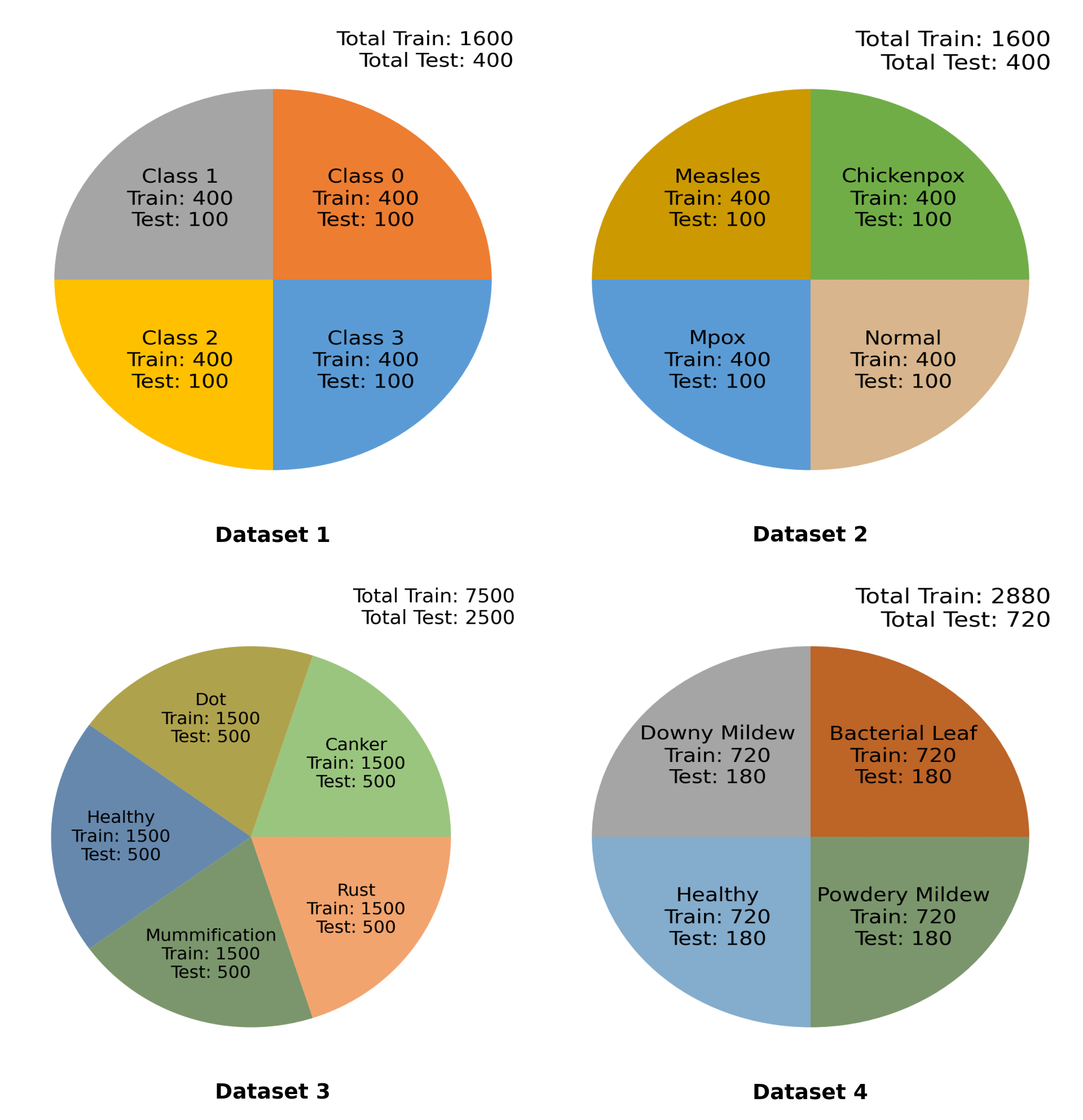} 
    \caption{Pie charts of dataset split and class distribution. All reported testing results and ablation analyses for all datasets were performed on their corresponding held-out test subsets.}
    \label{fig:dataset_samples}
\end{figure}

To demonstrate the adaptability of the proposed architecture across different domains and scales, we evaluated the model on four distinct datasets, ranging from synthetic patterns to real-world medical and agricultural images. A summary of the dataset splits and class distributions for all four datasets is illustrated in Fig.~\ref{fig:dataset_samples}.

\vspace{2mm}

\begin{figure}[htbp] 
    \centering
    \includegraphics[width=0.8\textwidth]{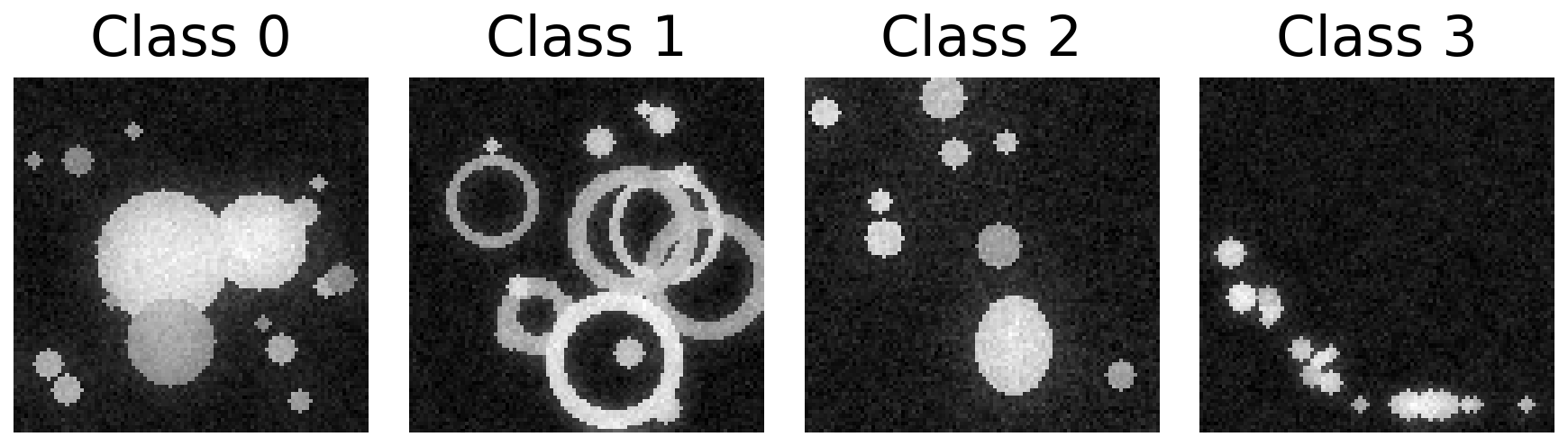} 
    \caption{Synthetic Spot Pattern Dataset (SSPD)}
    \label{fig:sspd}
\end{figure}

\noindent \textbf{(i) Dataset 1: Synthetic Spot Pattern Dataset (SSPD)} \\
The synthetic dataset contains four classes with different spatial structures, as shown in Fig.~\ref{fig:sspd}: (i) filled circular bulbs, (ii) hollow ring bulbs, (iii) elliptical bulbs with filament structures, and (iv) string-like bulbs arranged along a diagonal curve. All images were generated with randomly placed bulbs to simulate scattered, small, localized visual cues.

\vspace{2mm}

This dataset was specifically created to demonstrate the advantage of the fusion branch over the individual global and local branches. Component and parameter analyses were conducted only on real datasets to evaluate the effect of different model settings.

\noindent \textbf{(ii) Dataset 2: Skin Lesion Dataset} \\
The dataset contains 2,000 images distributed across four classes—chickenpox, measles, monkeypox, and normal. Class 1 (chickenpox) and class 2 (measles) share similar reddish lesion tones but differ in texture and spread: chickenpox lesions are small, raised, sparsely distributed blisters that appear in multiple stages across the body, while measles appears as flat or slightly raised red rashes with wider coverage. Class 3 (monkeypox) is distinctly different, usually showing large, raised pustules with sharp edges and deeper contrast. The normal class exhibits smooth and uniform skin texture. The dataset was created by combining public sources \cite{Bala2023} and \cite{Ali2022,Nafisa2024}. From MSLD v2.0, a four-class subset was selected, and data augmentation was applied to maintain class balance.

\vspace{2mm}

\noindent \textbf{(iii) Dataset 3: Guava Leaf Disease Dataset} \\
Guava leaf disease images were obtained from a publicly available dataset \cite{Guava} and augmented to increase sample size and balance class distributions. After augmentation, the dataset consisted of 10,000 images across five classes—canker, dot, mummification, rust, and healthy. This dataset was selected because the diseases primarily appear as visible spot patterns or localized regions on leaf surfaces, making them suitable for evaluating our model.

\vspace{2mm}

\noindent \textbf{(iv) Dataset 4: Grape Leaf Disease Dataset} \\
Grape leaf disease images were obtained from the publicly available NGLD dataset \cite{Dharrao2025a,NGLD_dataset} and augmented for class balance. After augmentation, the dataset consisted of 3,600 images across four classes: bacterial leaf spot, downy mildew, powdery mildew, and healthy. Similar to the guava dataset, these diseases exhibit spot-based infection patterns where affected regions vary in size, shape, and color intensity, while healthy leaves appear uniform.

\vspace{2mm}

\noindent\textbf{Data Splitting, Augmentation, and Duplicate Checking:}
For Datasets 2, 3, and 4, the original images were first divided into training and testing subsets. Augmentation was then applied separately to both subsets to maintain class balance. The augmentation operations included random horizontal flipping, rotation, zooming, width and height shifting, and image rescaling. To reduce the risk of data leakage, exact duplicates were identified using MD5 hash matching, while near-duplicate or visually similar images were detected using perceptual hashing (pHash) with a Hamming distance threshold of 5. Identified duplicate or visually similar images were removed from the test subset.

\subsection{Global Branch}

We selected DenseNet-121 as the backbone because it achieved the highest fusion accuracy across Datasets 2, 3, and 4, as shown in Table~\ref{tab:backbone_comparison}. The global branch uses this fine-tuned backbone to extract full-image contextual features, followed by two fully connected layers with \(L_2\) regularization, Batch Normalization, dropout of 0.4, and a final softmax classifier.

Training is performed for 15 epochs for dataset 1, 20 epochs for dataset 2, 15 epochs for dataset 3, and 20 epochs for dataset 4, using a batch size of 64, the Adam optimizer with an exponential-decay learning rate $1 \times 10^{-4}$ and categorical cross-entropy loss. Owing to differences in dataset scale, intra-class variability, and convergence behavior, the number of epochs was adjusted to balance learning progression and overfitting.

\begin{table}[h]
\centering
\caption{Performance comparison using multiple backbone architectures. DenseNet-121 delivered the best overall fusion accuracy across all three datasets.}
\label{tab:backbone_comparison}
\vspace{2mm}

\resizebox{\textwidth}{!}{%
\begin{tabular}{|l|c|c|c|c|c|c|c|c|c|}
\hline
 & \multicolumn{3}{c|}{\textbf{Dataset 2}} & \multicolumn{3}{c|}{\textbf{Dataset 3}} & \multicolumn{3}{c|}{\textbf{Dataset 4 (Grape)}} \\ \cline{2-10} 
\textbf{Backbone} & \textbf{Global} & \textbf{Local} & \textbf{Fusion} & \textbf{Global} & \textbf{Local} & \textbf{Fusion} & \textbf{Global} & \textbf{Local} & \textbf{Fusion} \\ \hline

\textbf{DenseNet-121} & 94.75\% & 94.25\% & 97.75\% & 98.76\% & 95.84\% & 99.64\% & 94.16\% & 89.44\% & 96.52\% \\ \hline

\textbf{DenseNet-169} & 95.75\% & 89.50\% & 91.75\% & 98.48\% & 94.84\% & 98.00\% & 92.92\% & 84.86\% & 93.89\% \\ \hline

\textbf{InceptionV3} & 96.25\% & 96.00\% & 97.00\% & 98.44\% & 93.32\% & 98.72\% & 93.33\% & 86.81\% & 92.64\% \\ \hline

\textbf{Xception} & 92.25\% & 91.25\% & 94.75\% & 99.00\% & 96.88\% & 99.20\% & 92.39\% & 88.75\% & 94.19\% \\ \hline

\end{tabular}%
}
\end{table}

\subsection{Attention Module}

\begin{figure}[h]
    \centering
    \includegraphics[width=1\textwidth]{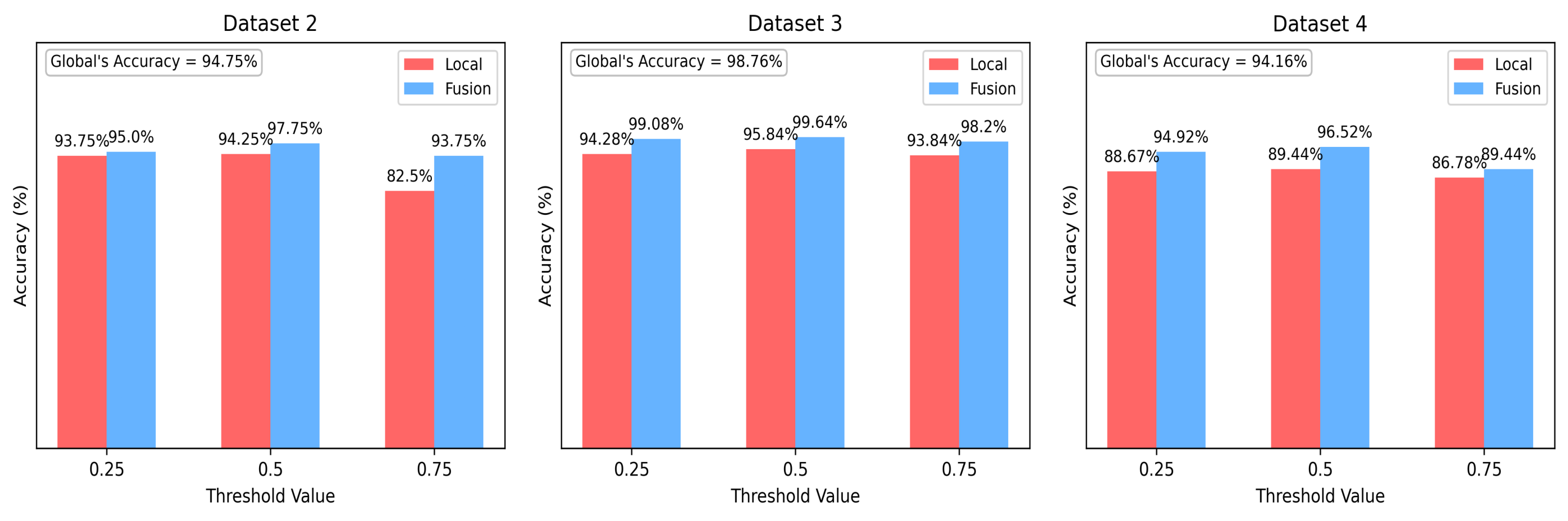} 
    \caption{Performance comparison across different Grad-CAM threshold values. Across all three datasets, a threshold of 0.5 consistently yielded the highest fusion accuracy. A lower threshold of 0.25 includes excessive background noise, whereas a higher threshold of 0.75 may mask out critical lesion features. Therefore, 0.5 was selected in this study, while the threshold remains adjustable based on dataset characteristics and performance.
    }
    \label{fig:threshold_comparison}
\end{figure}

\begin{figure}[h]
    \centering
    \includegraphics[width=1.0\textwidth]{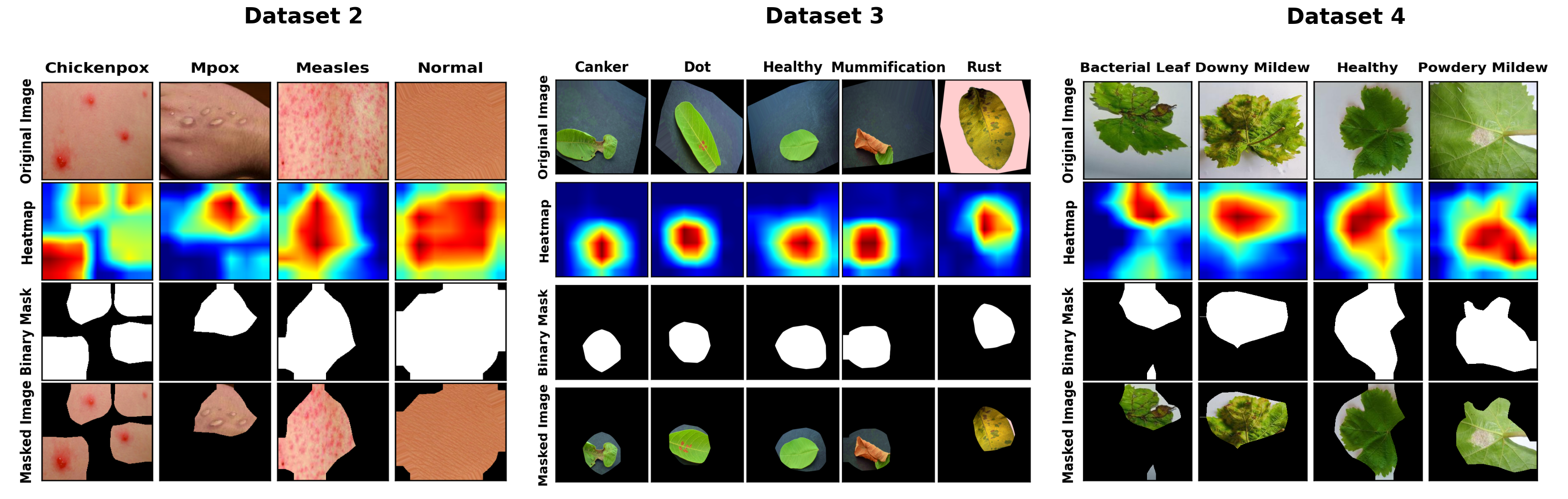} 
    \caption{Original images, Grad-CAM heatmaps, binary masks, and masked images for Datasets 2, 3, and 4.}
    \label{fig:imgs_d2}
\end{figure}

The final convolutional layer, \texttt{conv5\_block16\_concat}, was used to generate Grad-CAM heatmaps, which were normalized and thresholded at \(\tau=0.5\) (Fig.~\ref{fig:threshold_comparison}), to obtain binary masks. Figure~\ref{fig:imgs_d2} shows the original image, Grad-CAM heatmap, binary mask, and masked image used as input to the local branch.

\subsection{Local Branch}

\begin{figure}[htbp]
    \centering
    \includegraphics[width=0.6\textwidth]{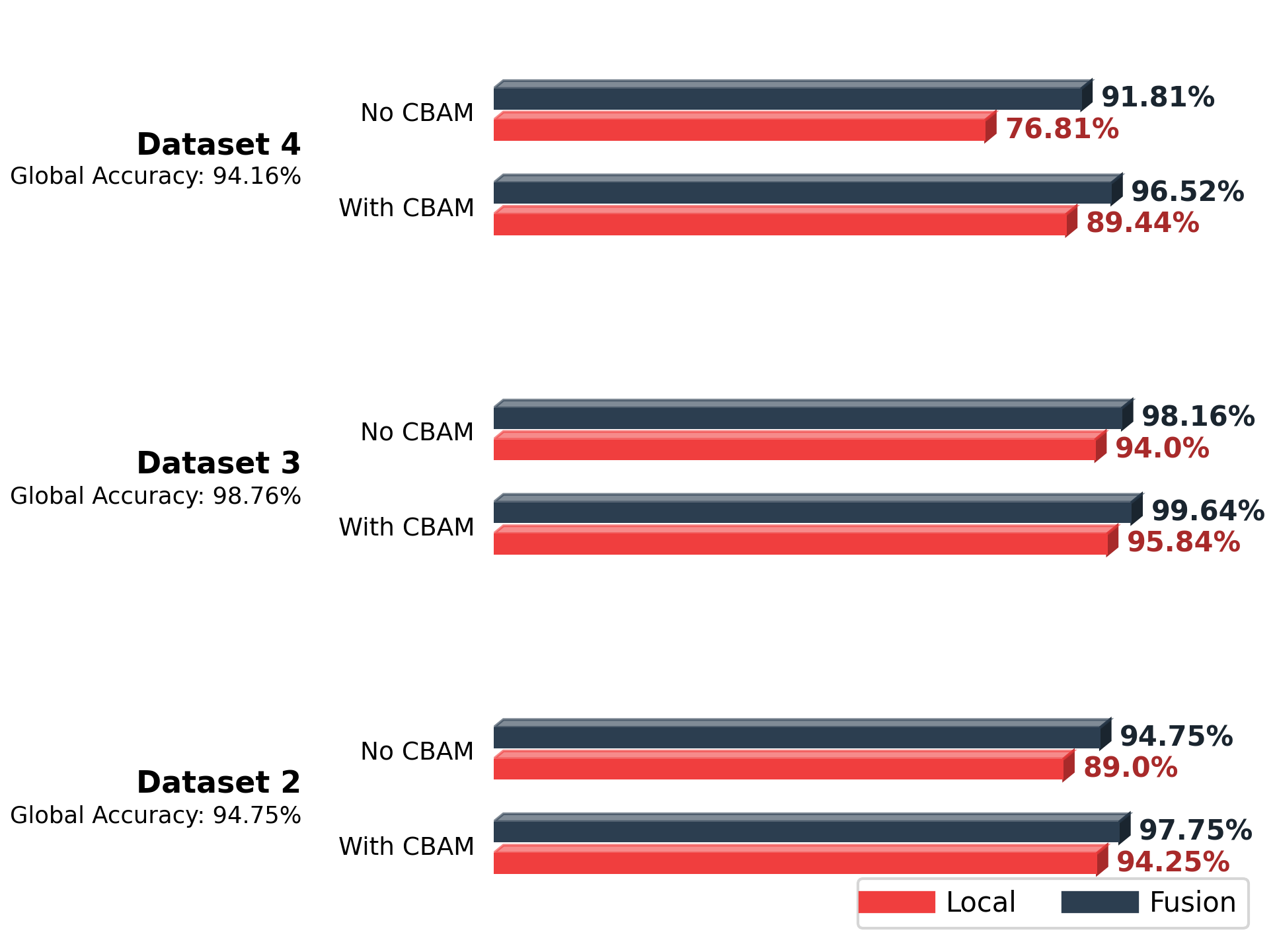} 
    \caption{Local branch with and without CBAM. Incorporating CBAM in the local branch consistently improves local and fusion accuracy across all datasets.}
    \label{fig:cbam_comparison}
\end{figure}

The local branch uses the same backbone architecture as the global branch but incorporates a Convolutional Block Attention Module (CBAM). As shown in Fig.~\ref{fig:cbam_comparison}, the CBAM-enhanced local branch outperformed the same architecture without CBAM, confirming its effectiveness for focused feature representation. Since the local model was trained on masked images, it used roughly half the number of epochs compared with the global branch, with a batch size of 16.

\begin{figure}[htbp]
    \centering
    \includegraphics[width=0.6\textwidth]{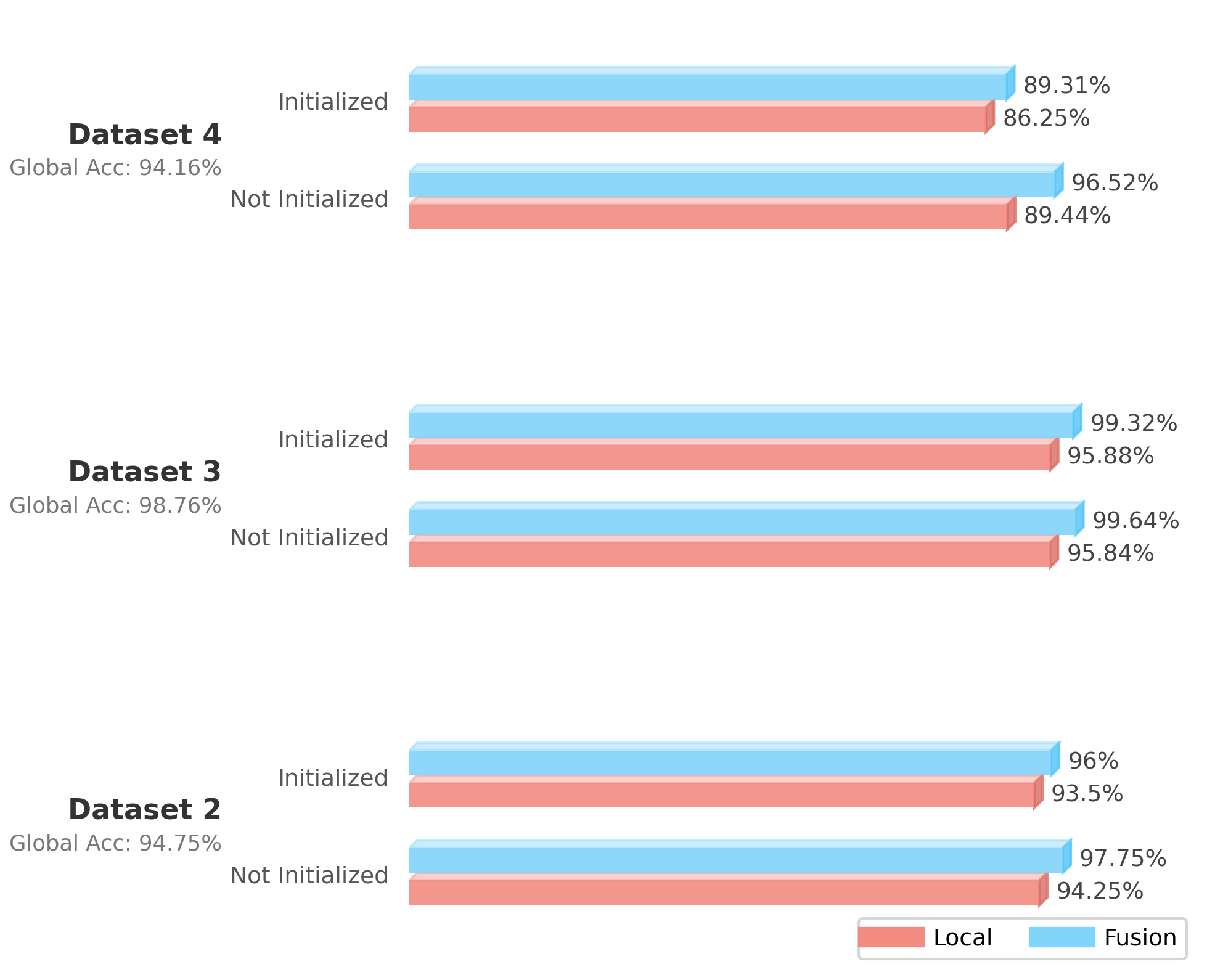} 
    \caption{Comparison of model performance with and without global weight initialization. While global weight initialization slightly improved the local branch accuracy for Dataset 3, fusion performance benefited more from independently trained branches. Both local and fusion performance degraded for Datasets 2 and 4 under global weight initialization.}
    \label{fig:weight_comparison}
\end{figure}

The local model was trained independently on masked images without global-weight initialization to avoid global-feature bias and better adapt to region-of-interest (ROI)-focused inputs. Although global-initialized weights improved local accuracy for dataset 3, independent training produced better fusion performance, as shown in Fig.~\ref{fig:weight_comparison}.

\subsection{Fusion Branch}

\begin{figure}[htbp]
	\centering
	\includegraphics[width=0.8\textwidth]{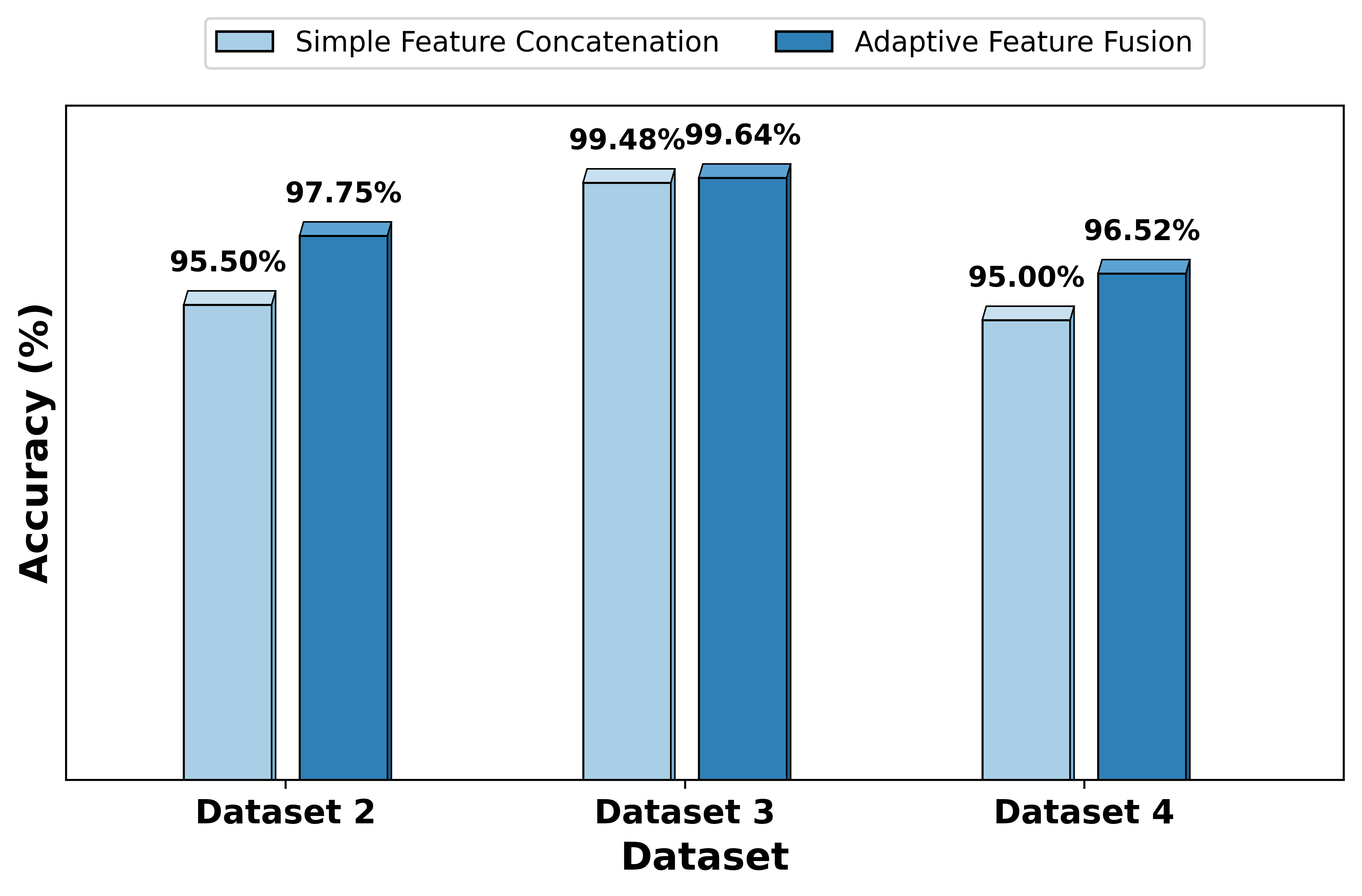} 
	\caption{Comparison between simple feature concatenation and adaptive feature fusion across Datasets 2, 3, and 4.
	}
	\label{fig:ada_fusion}
\end{figure}

\begin{figure}[htbp]
    \centering
    \includegraphics[width=0.9\textwidth]{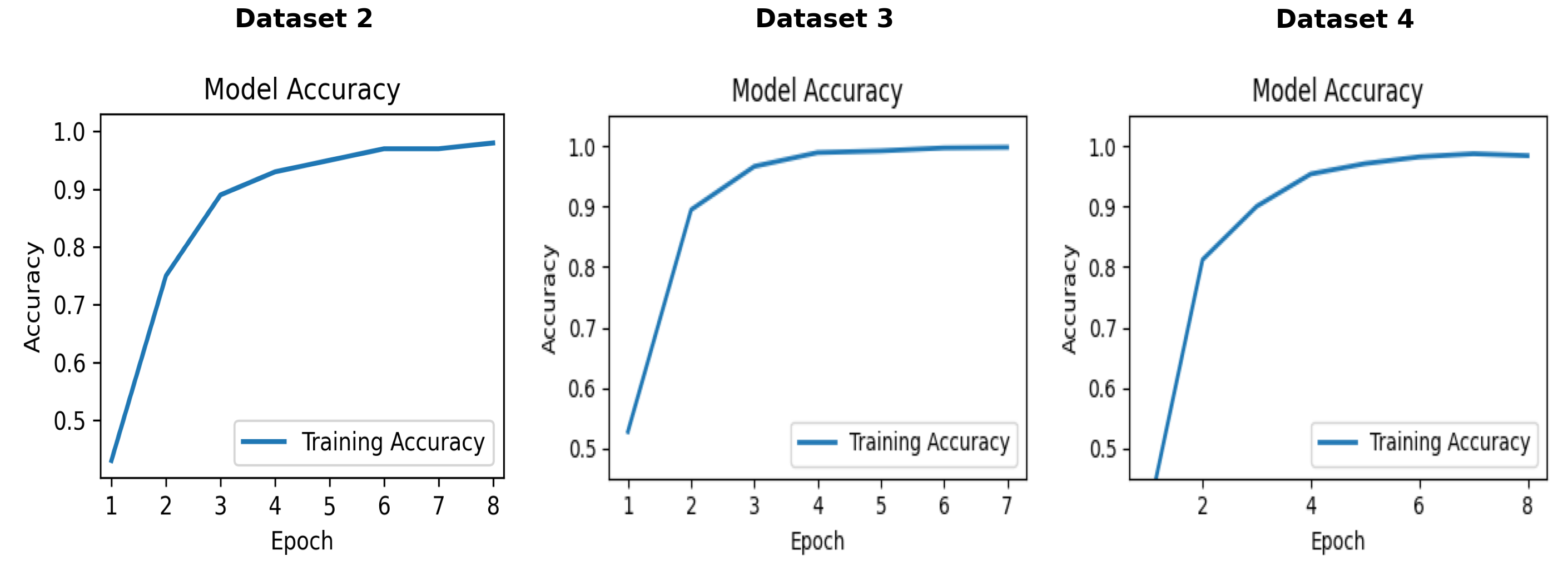} 
   \caption{Fusion model training accuracy curves for Datasets 2, 3, and 4}
    \label{fig:training_curve}
\end{figure}

The fusion branch was trained using the combined global and local feature representations with a batch size of 64 and the same optimizer and learning-rate schedule as the individual branches. Adaptive feature fusion was selected because it performed better than simple feature concatenation across Datasets 2, 3, and 4, as shown in Fig.~\ref{fig:ada_fusion}. 
The class-wise ranges of the learned fusion weights in Table~\ref{tab:fusion_weights} show substantial sample-level variation, indicating that the gating network dynamically adjusts global and local contributions rather than learning nearly constant weights.
The training accuracy curves of the fusion model are shown in Fig.~\ref{fig:training_curve}.

\begin{table}[htbp]
	\centering
	\caption{Class-wise ranges of learned global and local fusion weights on the Guava dataset.}
	\label{tab:fusion_weights}
	\begin{tabular}{lcc}
		\hline
		\textbf{Class} & 
		\textbf{$\alpha_g$ Min--Max} & 
		\textbf{$\alpha_l$ Min--Max} \\
		\hline
		Canker        & 0.183--0.898 & 0.102--0.817 \\
		Dot           & 0.256--0.938 & 0.062--0.744 \\
		Healthy       & 0.197--0.986 & 0.014--0.803 \\
		Mummification & 0.083--0.953 & 0.047--0.917 \\
		Rust          & 0.257--0.957 & 0.043--0.743 \\
		\hline
	\end{tabular}
\end{table}

\vspace{2mm}

\subsection{Result Analysis}

\vspace{4mm}

\begin{table}[h]
\centering
\caption{Performance comparison across four datasets}
\label{tab:performance_four_datasets}
\begin{adjustbox}{width=\textwidth}

\begin{tabular}{|c|c|c|c|c|c|c|c|c|c|c|c|c|}
\hline
\textbf{Metric} 
& \multicolumn{3}{c|}{\textbf{Dataset 1}}
& \multicolumn{3}{c|}{\textbf{Dataset 2}}
& \multicolumn{3}{c|}{\textbf{Dataset 3}}
& \multicolumn{3}{c|}{\textbf{Dataset 4}} \\ \hline

& \textbf{Global} & \textbf{Local} & \textbf{Fusion}
& \textbf{Global} & \textbf{Local} & \textbf{Fusion}
& \textbf{Global} & \textbf{Local} & \textbf{Fusion}
& \textbf{Global} & \textbf{Local} & \textbf{Fusion} \\ \hline

\textbf{Accuracy(\%)} 
& 97.50 & 93.75 & 98.75
& 94.75 & 94.25 & 97.75
& 98.76 & 95.84 & 99.64
& 94.16 & 89.44 & 96.52 \\ \hline

\textbf{F1-score} 
& 0.980 & 0.938 & 0.987
& 0.947 & 0.942 & 0.977
& 0.988 & 0.959 & 0.996
& 0.942 & 0.894 & 0.965 \\ \hline

\textbf{AUC (\%)} 
& 99.82 & 97.64 & 99.91
& 99.29 & 98.23 & 99.67
& 99.98 & 99.57 & 99.99
& 99.36 & 97.64 & 98.65 \\ \hline

\end{tabular}

\end{adjustbox}
\end{table}

Table~\ref{tab:performance_four_datasets} presents the overall accuracy, F1-score, and AUC for the Global, Local, and Fusion branches across all four datasets. 
A clear and consistent trend emerges across the datasets: the Fusion branch achieves the best performance in every case, demonstrating the advantage of integrating global contextual information with lesion-focused local representations.

For Dataset~1, the fusion branch achieves an accuracy gain of 1.25\% over the global branch and 5.00\% over the local branch. 
For Dataset~2, the improvements increase to 3\% and 3.5\%, respectively. 
Similarly, Dataset~3 shows gains of 0.88\% over the global model and 3.8\% over the local model. 
For Dataset~4, the fusion strategy improves accuracy by 2.36\% compared to the global branch and 7.08\% compared to the local branch.

Across these four datasets, the Local branch performed below the Global branch, indicating that Grad-CAM-based masking may suppress useful global structural cues such as shape, spatial distribution, and background texture that contribute to class separability. This suggests that the masked local input is not always sufficient when important class-related information exists outside the highlighted lesion region. The Fusion strategy compensates for this limitation by adaptively combining complementary feature representations, resulting in consistent gains in accuracy.

To further examine performance stability, experiments on Datasets 2 and 4 were repeated using four random seeds, as shown in Table~\ref{tab:seed_accuracy}. The Fusion branch showed stable performance and consistently achieved higher accuracy than the Global and Local branches.

\begin{table}[htbp]
\centering
\caption{Seed-wise accuracy comparison for Datasets 2 and 4}
\label{tab:seed_accuracy}
\renewcommand{\arraystretch}{1.2}
\setlength{\tabcolsep}{5pt}
\begin{tabular}{llccccc}
\hline
\textbf{Dataset} & \textbf{Branch} & \textbf{Seed 7} & \textbf{Seed 42} & \textbf{Seed 99} & \textbf{Seed 123} & \textbf{Mean $\pm$ SD (\%)} \\
\hline

\multirow{3}{*}{Skin lesion} 
& Global & 95.50 & 94.75 & 96.00 & 94.25 & 95.13 $\pm$ 0.78 \\
& Local  & 94.25 & 94.75 & 95.75 & 93.25 & 94.50 $\pm$ 1.04 \\
& Fusion & 98.00 & 97.25 & 97.75 & 97.25 & \textbf{97.56 $\pm$ 0.38} \\

\hline

\multirow{3}{*}{Grape dataset} 
& Global & 94.86 & 94.02 & 94.44 & 94.72 & 94.51 $\pm$ 0.37 \\
& Local  & 88.19 & 89.44 & 89.10 & 90.13 & 89.22 $\pm$ 0.81 \\
& Fusion & 95.83 & 96.11 & 95.97 & 96.66 & \textbf{96.14 $\pm$ 0.36} \\

\hline
\end{tabular}
\end{table}

\begin{figure}[!htbp]
\centering

\begin{subfigure}{0.46\textwidth}
    \centering
    \includegraphics[width=\linewidth]{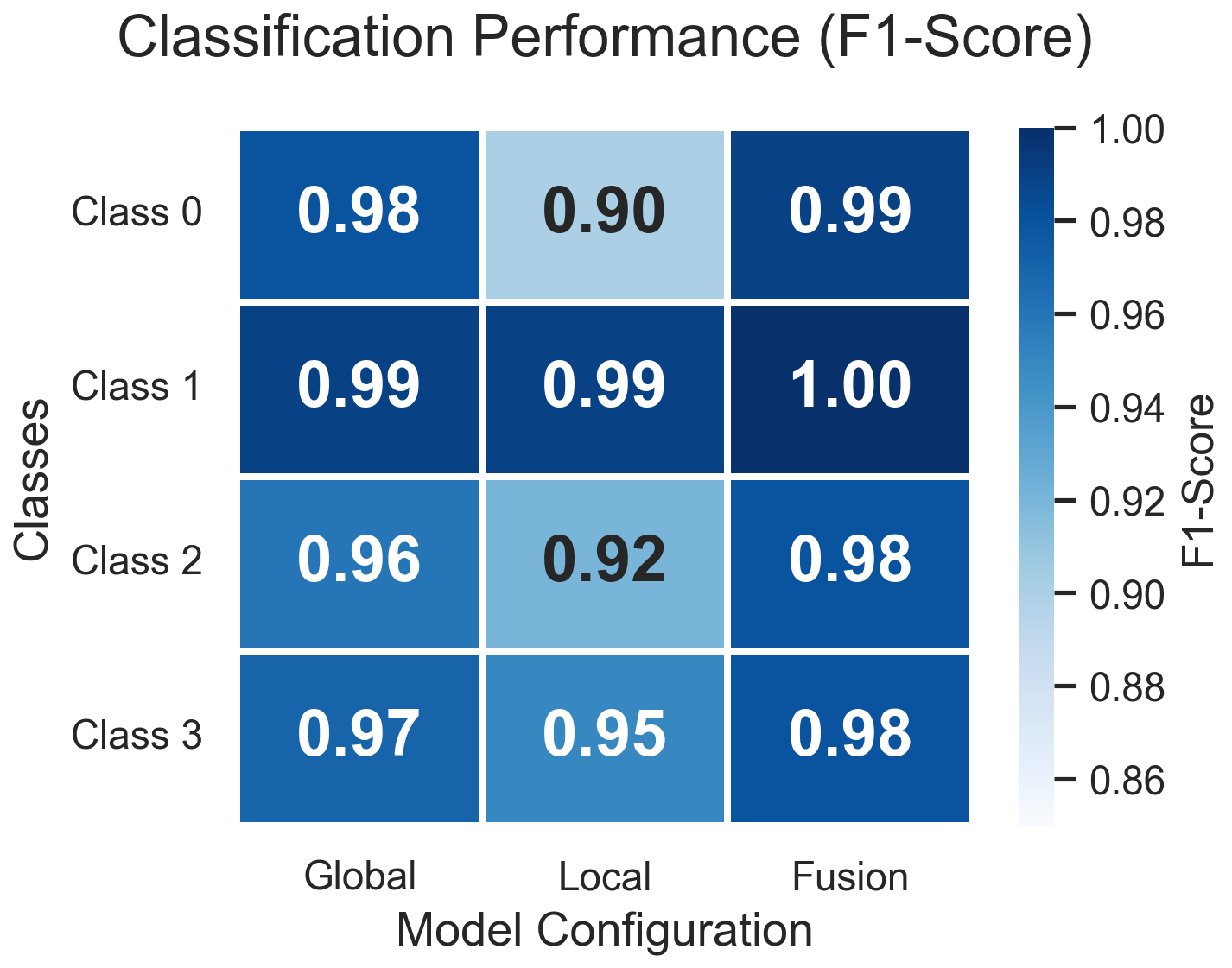}
    \caption{Dataset 1}
\end{subfigure}
\hfill
\begin{subfigure}{0.46\textwidth}
    \centering
    \includegraphics[width=\linewidth]{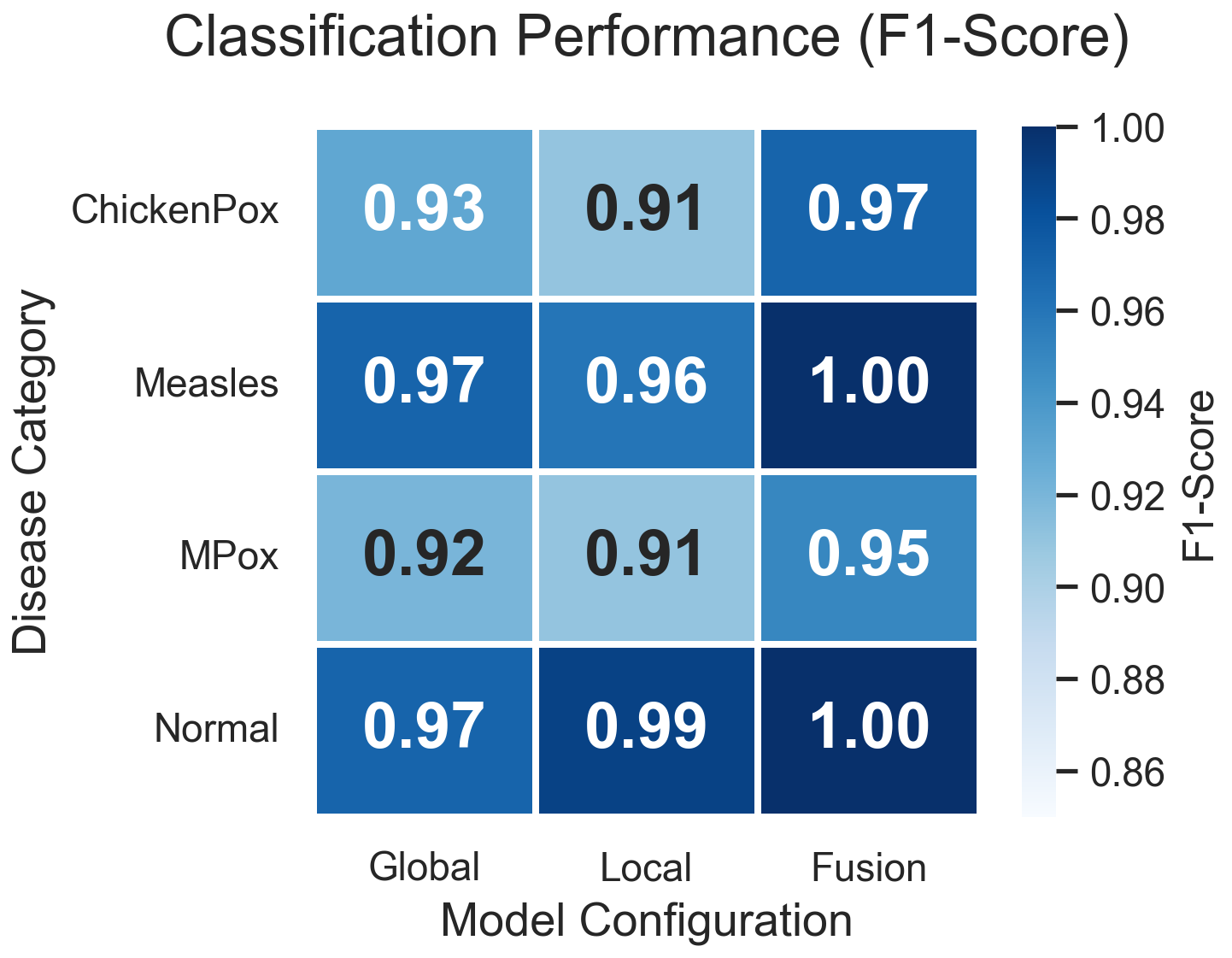}
    \caption{Dataset 2}
\end{subfigure}

\vspace{0.5cm}

\begin{subfigure}{0.46\textwidth}
    \centering
    \includegraphics[width=\linewidth]{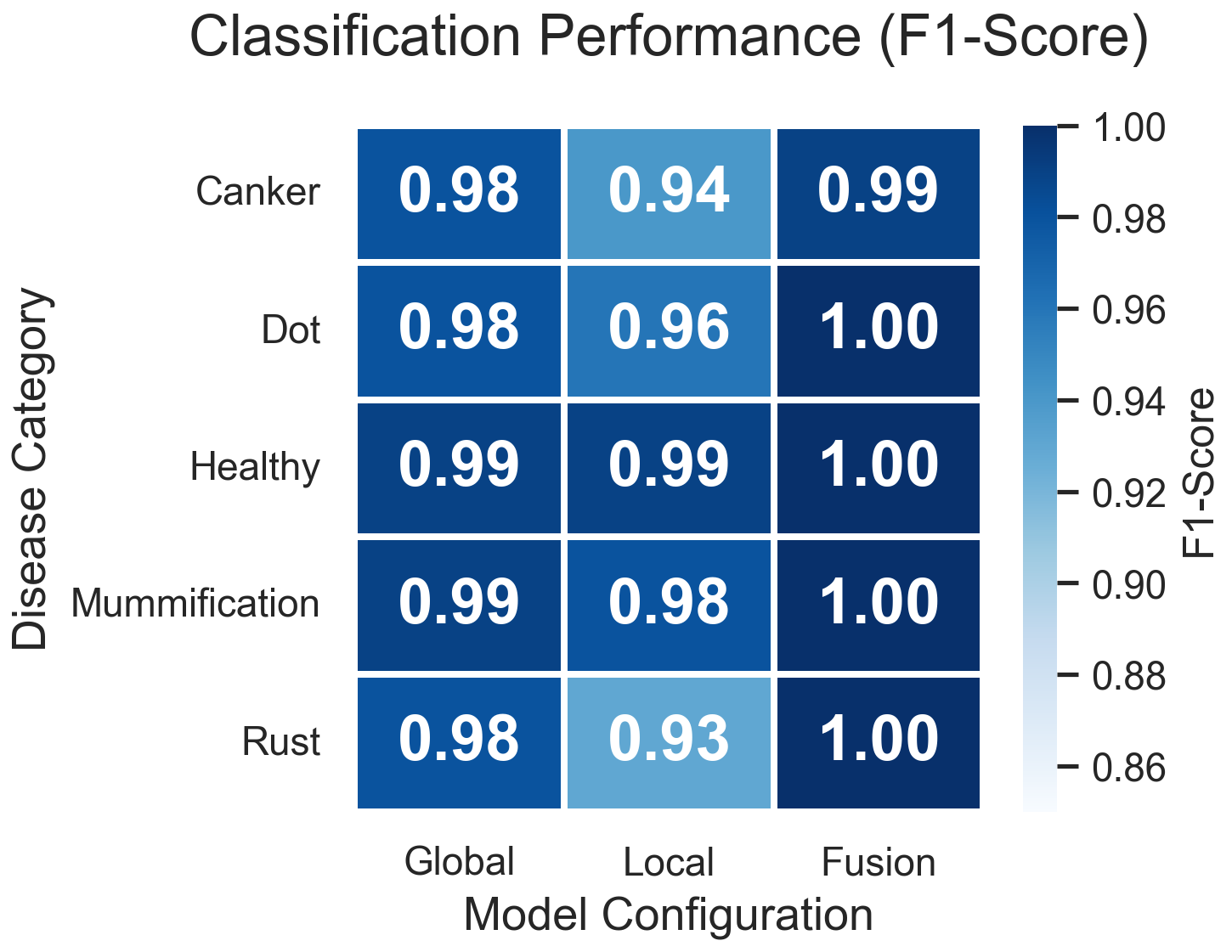}
    \caption{Dataset 3}
\end{subfigure}
\hfill
\begin{subfigure}{0.46\textwidth}
    \centering
    \includegraphics[width=\linewidth]{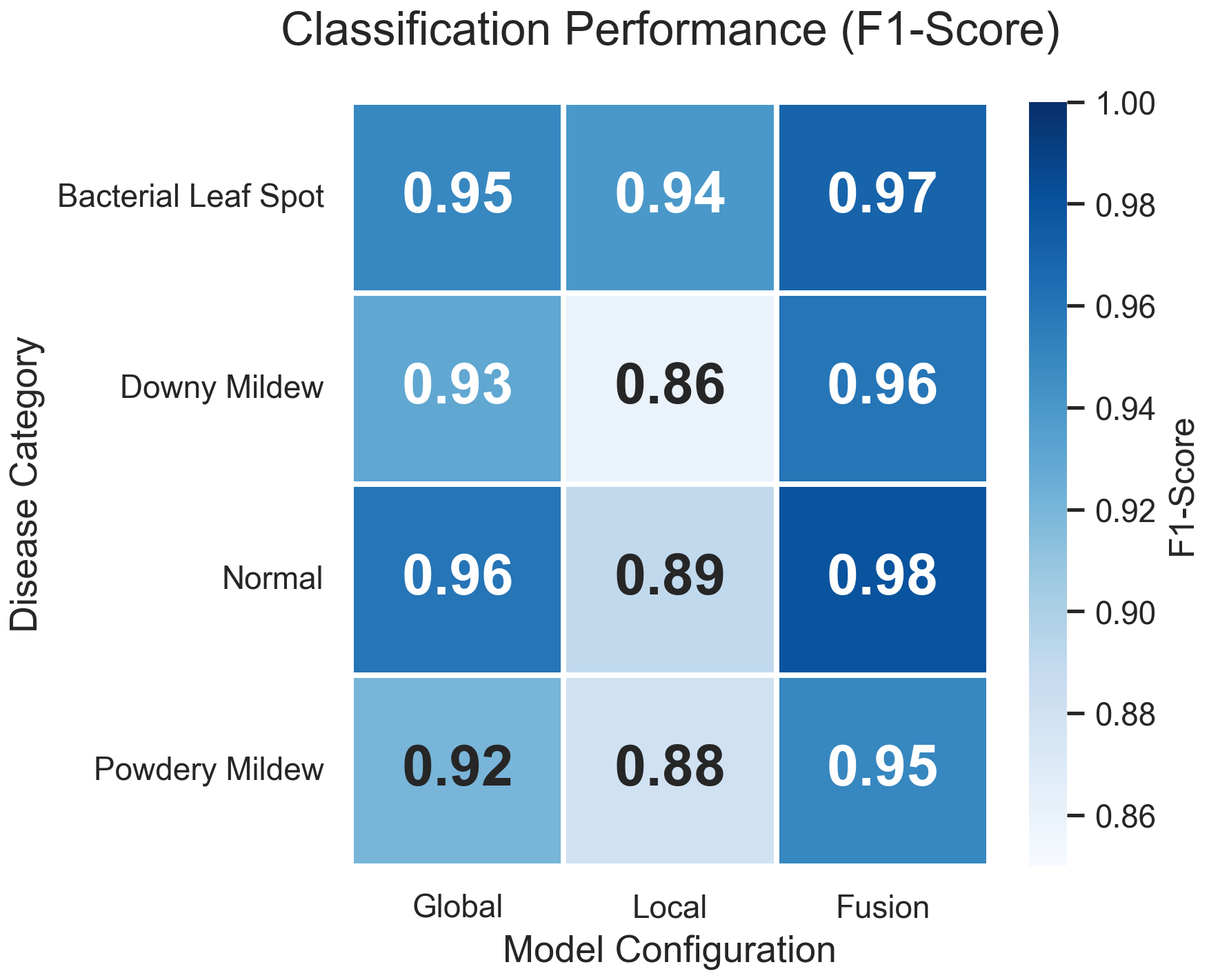}
    \caption{Dataset 4}
\end{subfigure}

\caption{Class-wise F1-score for all four datasets}
\label{fig:all_f1_scores}

\end{figure}

The class-wise F1-score analysis (Fig.~\ref{fig:all_f1_scores}) further supports the overall trend. In Dataset 1, the Local branch shows lower F1-scores for classes 0, 2, and 3, while the Fusion branch improves performance across all classes. In Dataset 2, the Local branch improves the F1-score for the normal class but reduces it for chickenpox, measles, and monkeypox; however, the Fusion branch improves the F1-scores across all four classes. Dataset 3, where the Local branch shows comparatively lower F1-scores for all classes except healthy, the Fusion branch consistently achieves the highest class-wise performance. In Dataset 4, the Local branch experiences the most noticeable degradation in multiple classes; however, the Fusion configuration restores strong and balanced F1-scores, surpassing both individual branches.

\begin{figure}[htbp]
\centering

\includegraphics[width=0.9\textwidth]{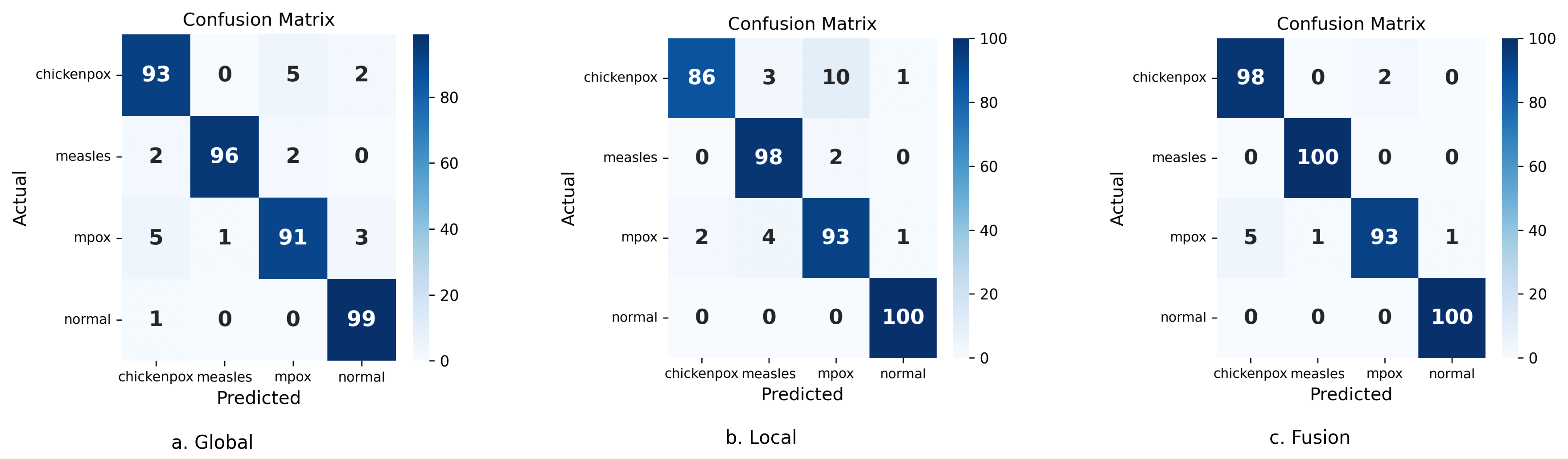}
\caption*{(a) Dataset 2}

\vspace{0.5cm}

\includegraphics[width=0.9\textwidth]{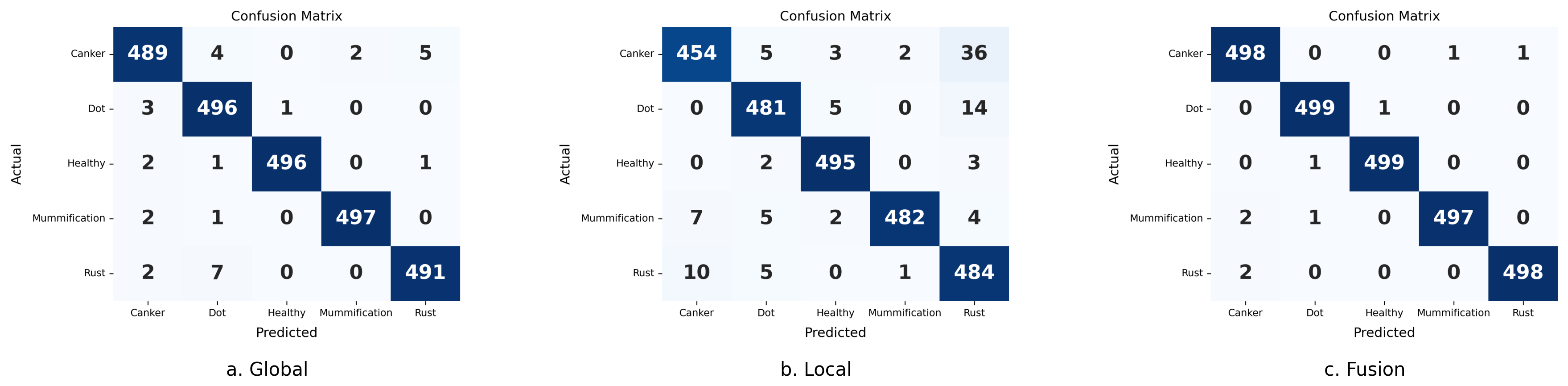}
\caption*{(b) Dataset 3}

\caption{Confusion matrices for Datasets 2 and 3}
\label{fig:confusion_matrices}

\end{figure}

From the confusion matrices (Fig.~\ref{fig:confusion_matrices}), for Dataset 2, the global model's errors occurred across all classes. Seven chickenpox samples were misclassified. Monkeypox showed confusion with chickenpox (5 cases), measles (1 case), and normal (3 cases). In the Local branch, errors were reduced for measles, monkeypox, and normal, but increased for chickenpox. When the Fusion model was applied, these cross-class errors were further reduced, with a total of nine misclassifications.

In Dataset 3, the Global branch performs strongly, with only limited confusion among related leaf diseases. However, the Local branch shows a large rise in misclassification toward the rust class. Specifically, 36 canker samples and 14 dot samples were predicted as rust, while 10 rust samples were predicted as canker. The Fusion branch substantially reduces this error. Correct predictions increased across all classes: 498 canker, 499 dot, 499 healthy, 497 mummification, and 498 rust.

\subsection{Paired Statistical Comparison}

Table~\ref{tab:mcnemar_guava} presents the paired comparison between the Global and Fusion models on the Guava dataset. The Global model misclassified 31 samples, whereas the Fusion model misclassified 9. Among the paired predictions, the Fusion model corrected 26 errors made by the Global model, while 4 samples correctly classified by the Global model were misclassified by the Fusion model; five samples were misclassified by both models. The exact McNemar test showed a statistically significant difference ($p = 5.95 \times 10^{-5}$), supporting the performance improvement achieved by the Fusion model over the Global model.

\begin{table}[htbp]
	\centering
	\caption{Paired statistical comparison of the Global and Fusion models on the Guava dataset.}
	\label{tab:mcnemar_guava}
	\begin{tabular}{lc}
		\hline
		\textbf{Measure} & \textbf{Guava Dataset} \\
		\hline
		Global errors & 31 \\
		Fusion errors & 9 \\
		Wrong by both models & 5 \\
		Global wrong $\rightarrow$ Fusion correct & 26 \\
		Global correct $\rightarrow$ Fusion wrong & 4 \\
		McNemar $p$-value & $\mathbf{5.95 \times 10^{-5}}$ \\
		\hline
	\end{tabular}
\end{table}

\subsection{Source-Held-out Experiment}

\begin{table}[htbp]
\centering
\caption{Source-held-out accuracy on the MSID dataset}
\label{tab:source_held_out_accuracy}
\renewcommand{\arraystretch}{1.2}
\begin{tabular}{lc}
\hline
\textbf{Model Branch} & \textbf{Accuracy (\%)} \\
\hline
Global & 79.11 \\
Local & 82.55 \\
Fusion & \textbf{83.83} \\
\hline
\end{tabular}
\end{table}

We performed a source-held-out experiment to evaluate the model under a cross-source testing setting. The model was trained on the MSLD v2.0 dataset, where augmentation was applied only to the training images, and then tested on the original MSID dataset without augmentation. The test images were kept unaugmented to evaluate performance on raw, unseen data. Before testing, exact and near-duplicate images between the training and test sets were removed to reduce the risk of possible data leakage.

As shown in Table~\ref{tab:source_held_out_accuracy}, the Global, Local, and Fusion branches achieved accuracies of 79.11\%, 82.55\%, and 83.83\%, respectively. Unlike the main experimental datasets, the Local branch performed better than the Global branch. This may be because the global appearance and background patterns vary more between MSLD v2.0 and MSID, while lesion-focused local features are relatively more transferable across sources. The Fusion branch achieved the highest accuracy by combining both global context and local lesion-specific information.

\begin{figure}[htbp]
    \centering
    \includegraphics[width=0.6\textwidth]{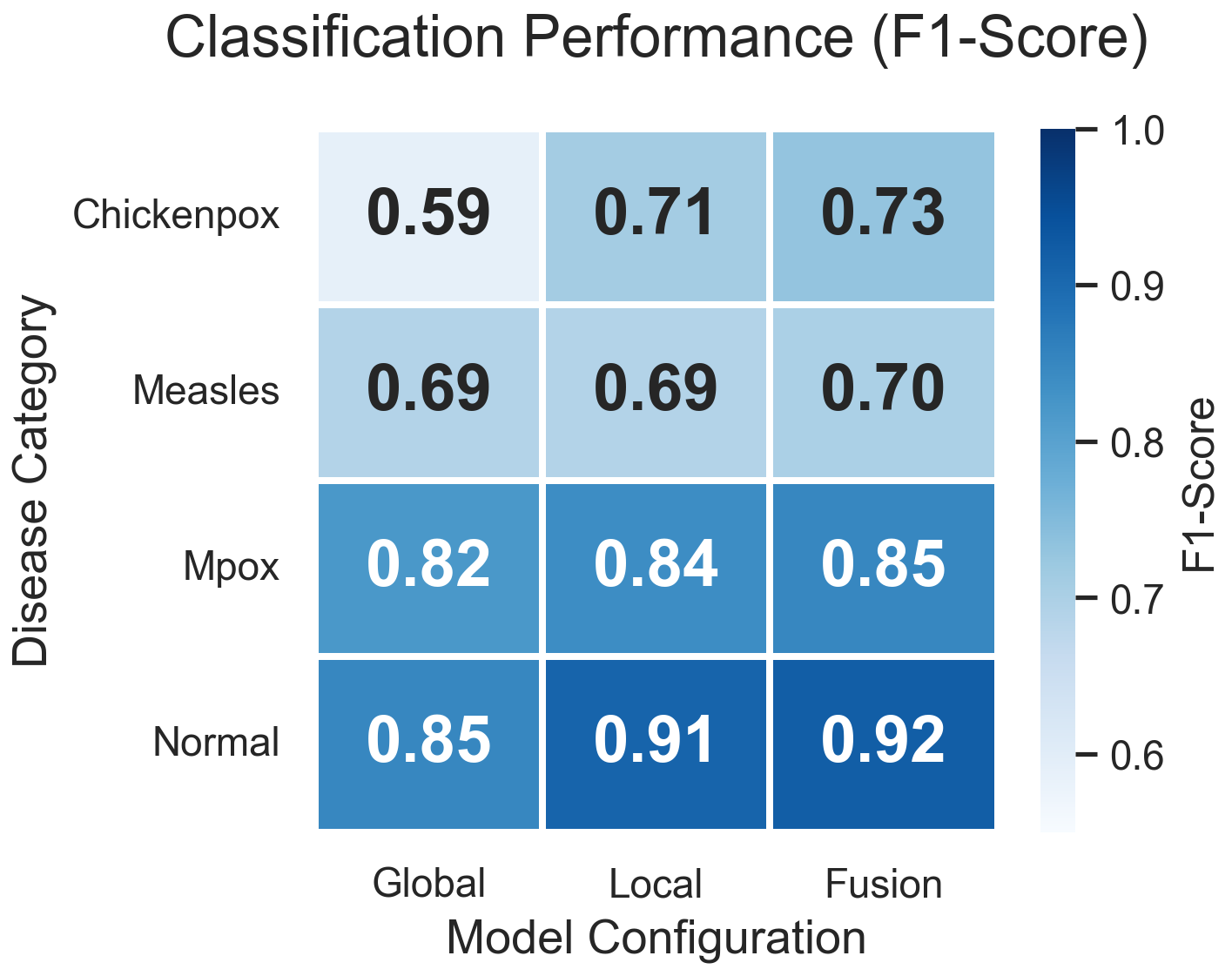} 
    \caption{Class-wise F1-score comparison of the global, local, and fusion branches in the source-held-out experiment.}
    \label{fig:f1_source}
\end{figure}

The class-wise F1-scores (Fig.~\ref{fig:f1_source}) further show stronger performance for monkeypox and normal class, while chickenpox and measles achieved lower scores. This may be due to the visual similarity between lesion classes, source-related domain shift, and the smaller number of test samples for some classes. Compared with the within-dataset result for Dataset 2, the lower source-held-out accuracy also shows that cross-source testing remains more challenging. This performance drop may be related to differences in image acquisition, background appearance, lesion presentation, and class distribution across sources. Overall, the Fusion branch still achieved the highest accuracy among the three branches, suggesting that the proposed fusion strategy provides a better balance between global context and lesion-focused local features under cross-source variation. 

\subsection{Performance Comparison with Existing Studies}

To assess the effectiveness of the proposed global--local--fusion framework, a contextual comparison with existing studies on skin lesion, guava leaf, and grape leaf disease datasets is presented in Table~\ref{tab:overall_performance}.

The acronyms used in Table~\ref{tab:overall_performance} are defined as follows: LIME, Local Interpretable Model-Agnostic Explanations; ViT, Vision Transformer; MAE, Masked Autoencoder; DINO, self-Distillation with No Labels; KNN, k-nearest neighbors; SVM, support vector machine; ANN, artificial neural network; and GAN, generative adversarial network.

\renewcommand{\arraystretch}{1.2}
\setlength{\tabcolsep}{3pt}
\scriptsize

\begin{longtable}{p{1.8cm} p{1.6cm} p{2.0cm} p{1.1cm} p{4.6cm} p{1.2cm}}

\caption{Contextual performance comparison with existing studies on skin lesion, guava leaf, and grape leaf disease datasets.}
\label{tab:overall_performance} \\

\hline
\textbf{Application Domain} & 
\textbf{Paper} & 
\textbf{Dataset} & 
\textbf{Classes} & 
\textbf{  Method} & 
\textbf{Accuracy} \\
\hline
\endfirsthead

\hline
\textbf{Application Domain} & 
\textbf{Paper} & 
\textbf{Dataset} & 
\textbf{Classes} & 
\textbf{  Method} & 
\textbf{Accuracy} \\
\hline
\endhead

\hline
\endfoot


Skin Lesion 
& \cite{SorayaieAzar2023a} 
& Public dataset 
& 2,4 
& Developed multiple deep neural network models, with a modified DenseNet201-based architecture as the main model. LIME and Grad-CAM were incorporated to provide visual explainability. 
& 97.63\% (2-class), 95.18\% (4-class) \\

& \cite{Ahmed2025} 
& MCVSLD dataset
& 6 
& Proposed a modified Xception-based deep learning architecture
& 96.36\%, 97.01\% \\

& \cite{Vuran2025a} 
& MSLD v2.0 
& 6 
& Evaluated transformer-based deep learning architectures, including ViT, MAE, DINO, and Swin Transformer. The study investigated the effects of self-supervised learning, self-distillation, and shifted-window techniques on classification performance
& 93.71\% \\

& \cite{KumarSaha2025a} 
& MSID 
& 4 
& Proposed Mpox-XDE, an ensemble architecture combining modified Xception, DenseNet201, and EfficientNetB7 models for classification, with Grad-CAM applied for explainable visualization. 
& 98.70\% \\

& \cite{Arafa2025a} 
& MSLD, MSID 
& 2,4 
& Introduced MSCADMpox, a multi-stage classification framework combining generative augmentation, handcrafted feature extraction, and deep feature extraction. The method used ViT, VGG16, VGG19, ResNet50, MBGWO-based feature selection, and traditional classifiers for final decision-making. 
& 90.54\% (MSLD), 91.88\% (MSID) \\

& \textbf{Our Model} 
& MSLD v2.0, MSID 
& 4 
& Proposed global--local--fusion framework combining global image features, local lesion-focused features, and adaptive fusion for skin lesion classification 
& \textbf{97.75\%} \\

\hline


Guava Leaf 
& \cite{Doutoum2023a} 
& Publicly available dataset 
& 5 
& Applied pretrained convolutional neural networks for automated guava leaf disease classification. The method extracted visual features from leaf images and classified disease categories using deep CNN-based classifiers. 
& 94.93\% \\

& \cite{Thangaraj2023a} 
& Kaggle dataset
& 5 
& Employed transfer learning with pretrained CNN architectures, including DenseNet121, DenseNet169, InceptionV3, and Xception. The study compared these backbones to identify the most effective model for extracting disease-discriminative leaf features.
& 96.12\% \\

& \cite{MustakUnNobi2023a} 
& Publicly available dataset (Dataset 1) 
& 5 
& Introduced GLD-Det, a lightweight deep learning model based on a modified MobileNet architecture. The model integrates pooling layers, batch normalization, dropout, dense layers, and a softmax classifier for real-time guava leaf disease detection. 
& 98.00\% \\

& \cite{MuhammadAsim2025a} 
& Public + self-acquired images 
& 8 
& Proposed a computational intelligence-based ensemble framework combining KNN, SVM, ANN, and Random Forest classifiers. The outputs of these models were fused through a meta-learning strategy to improve disease classification robustness.
& 97.32\% \\

& \cite{Guler2025a} 
& Publicly available dataset 
& 5 
& Proposed a hybrid multi-channel deep learning framework combining traditional augmentation, GAN-based synthetic image generation, and ensemble learning. The architecture fused complementary features from InceptionV3 and ResNet50 for guava leaf disease classification.
& 97.50\% \\

& \textbf{Our Model} 
& Kaggle dataset 
& 5 
& Proposed global--local--fusion framework for guava leaf disease classification using complementary global and local disease-related features 
& \textbf{99.64\%} \\

\hline


Grape Leaf 
& \cite{Eren2025a} 
& NGLD 
& 4 
& Employed a Swin Transformer Tiny architecture for grape leaf disease classification. The model used shifted-window self-attention to capture both local disease symptoms and broader visual patterns from leaf images. 
& 98.80\% \\

& \cite{Dharrao2025a} 
& NGLD 
& 4 
& Introduced the Niphad Grape Leaf Disease Dataset (NGLD). Validated the dataset via a transfer learning approach using ResNet-18 architecture. 
& 96.00\% \\

& \cite{Loganathan2025a}
& PlantVillage (PV) dataset, NGLD
& 4
& Proposed ELCAM-Net, an automated lightweight channel attention-based CNN integrated with explainable AI. The framework incorporates a global--local context-enhanced channel attention module into the SqueezeNet backbone to refine feature-map channel weights, suppress irrelevant information, and emphasize disease-specific leaf patterns.
& 99.41\% (PlantVillage dataset), 93.4\% (NGLD) \\

& \textbf{Our Model} 
& NGLD 
& 4 
& Proposed global--local--fusion framework for grape leaf disease classification using integrated global context and local disease-specific feature learning 
& \textbf{96.52\%} \\

\hline

\end{longtable}

\normalsize

\section{Discussion}\label{sec5}
The experimental results across four datasets consistently demonstrate that lesion-focused classification benefits from jointly modeling global contextual information and localized discriminative regions. While the global and local branches capture complementary features, adaptive fusion combines them more effectively, leading to more stable and accurate predictions.

Compared with existing studies in Table~\ref{tab:overall_performance}, the proposed framework can be positioned within the broader research context. Several studies have achieved strong results using single-backbone convolutional neural network (CNN) classifiers~\cite{SorayaieAzar2023a,Ahmed2025,Doutoum2023a,Thangaraj2023a,MustakUnNobi2023a}. Other studies improved performance using fusion or multi-stage strategies. Some combined multiple CNN backbones or classical classifiers for stronger final prediction~\cite{KumarSaha2025a,MuhammadAsim2025a}. Arafa et al.~\cite{Arafa2025a} used a multi-stage pipeline with synthetic image generation, handcrafted and deep feature extraction, feature selection, and traditional classifiers, while Guler et al.~\cite{Guler2025a} fused features from models trained on GAN-generated synthetic images and traditionally augmented images using a stacking ensemble. These findings support the importance of stronger feature extraction and fusion for disease classification. Our work extends this direction by forming two complementary representations of each sample: the original full image for global contextual learning and the Grad-CAM-masked image for lesion-focused local learning. In contrast to studies in which Grad-CAM is mainly used after prediction for explainability, the proposed framework uses Grad-CAM within the learning pipeline to guide local feature extraction. The proposed findings are also consistent with transformer-based and attention-based studies that emphasize the importance of capturing localized disease patterns and broader contextual information~\cite{Vuran2025a,Eren2025a,Loganathan2025a}.

Throughout our experiments, a clear pattern was observed: the global branch provided strong baseline performance, particularly when overall structure, color distribution, and background information helped distinguish the classes. However, it may still confuse classes with similar global appearance but different lesion details. 

The local branch focuses more effectively on lesion-specific details but often at the cost of losing contextual structure. This trade-off is evident in all four datasets, where removing background information increases inter-class confusion. However, the source-held-out experiment shows a different behavior, where the local branch performs better than the global branch, indicating that lesion-centered features may be more transferable when images come from different sources.

The fusion branch consistently achieved the highest accuracy by dynamically balancing these complementary representations. Rather than assuming equal contribution from global and local features, the adaptive weighting mechanism allows the model to emphasize the most reliable source of information for each image. This behavior is especially beneficial in datasets where lesion visibility, size, and distribution vary widely. For images with subtle or sparse lesions, the model can rely more on global context, whereas for images with strong localized disease patterns, the local branch receives higher importance. This sample-specific adaptability explains the consistent reduction in confusion errors observed in the fusion confusion matrices across all datasets.

Another important observation is the role of attention guidance quality. The Grad-CAM masks generated by the global branch generally align well with discriminative regions, enabling the local branch to learn meaningful lesion-centric features. However, performance drops in the local branch suggest that Grad-CAM localization, while helpful, is not perfect and may occasionally omit contextual cues that remain important for classification. This further justifies the need for fusion rather than relying solely on ROI-based learning.

The ablation studies also highlight that component-level design choices significantly influence performance. Incorporating CBAM improves the discriminative ability of the local branch by refining channel and spatial attention within masked regions. Similarly, independent training of global and local branches leads to better fusion performance than weight sharing, suggesting that representation diversity between branches enhances complementary learning. Overall, the results indicate that optimal performance arises from adaptive cooperation between global context understanding and fine-grained lesion analysis, which is what the proposed fusion framework is designed to achieve.

\subsection{Limitations and Future Study}
\label{sec:results}

\noindent \textbf{Sensitivity to Background Complexity}

\begin{figure}[t]
    \centering
    \includegraphics[width=0.5\textwidth]{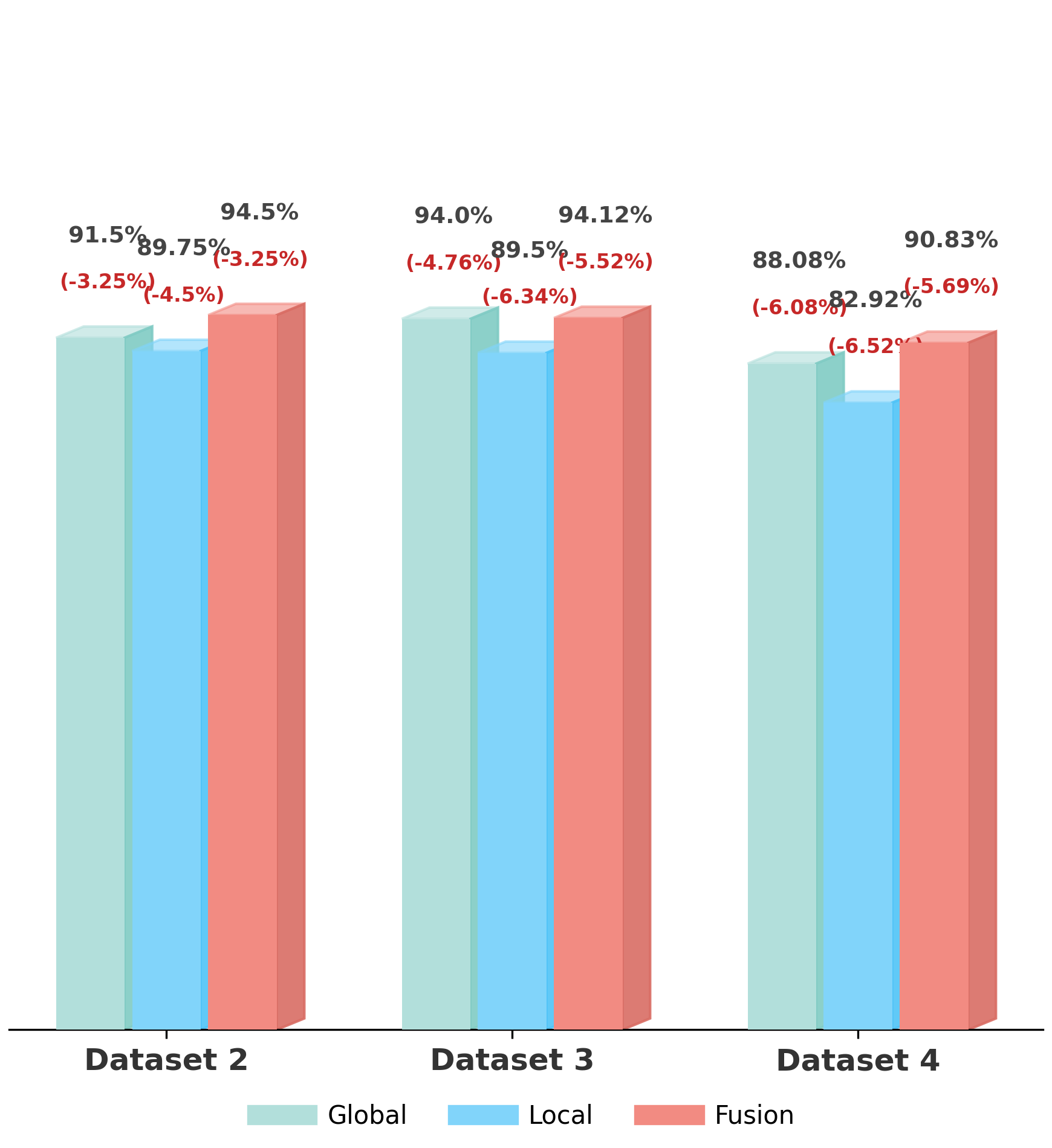} 
    
    \caption{Comparative performance of the global--local--fusion branches under increased background complexity using Gaussian noise ($\sigma=70.0$).}
    
    \label{fig:noise_comparison}
\end{figure}

The proposed architecture performs most effectively when the subject is clear and background clutter is limited, allowing Grad-CAM to produce more focused attribution maps. However, its performance degrades as background complexity increases. To systematically evaluate this effect, Gaussian noise was added to non-ROI regions while preserving the lesion area and a 10-pixel safety margin. As shown in Fig.~\ref{fig:noise_comparison}, the performance of the global, local, and fusion branches decreases under increased background noise, indicating that complex backgrounds remain a limitation. 
\vspace{0.17cm}
This degradation is mainly due to the sensitivity of the global branch to background variations, which can affect Grad-CAM mask quality and subsequently influence the local branch. Addressing this limitation motivates future work on incorporating more robust preprocessing strategies and developing noise-aware attention mechanisms to improve resilience under complex background conditions.

Another important limitation is that the local branch depends on the quality of Grad-CAM-based masks, where non-highlighted regions are suppressed. Although this helps the model focus on lesion-specific regions, it may also remove useful contextual information such as lesion shape, spatial distribution, surrounding texture, and background patterns that contribute to class discrimination. This explains why the local branch sometimes performs worse than the global branch in the experimental results. The fusion branch mitigates this limitation by combining global and local features; however, improving ROI localization remains an important direction for future work.

\subsection{Theoretical and Practical Implications}

The findings of this study provide important theoretical insights into lesion-focused image classification. The results demonstrate that global contextual features and local lesion-specific features provide complementary information, and neither feature source alone is sufficient for consistently optimal performance. The improved performance of the fusion branch confirms the importance of adaptively balancing global and local representations. This supports the theoretical idea that lesion-focused classification should not depend exclusively on either full-image learning or ROI-based learning. Instead, combining Grad-CAM-guided localization, CBAM-based local refinement, and adaptive sample-wise fusion can provide a stronger framework for learning discriminative features from images.

From a practical perspective, the proposed framework is useful for lesion- or spot-based image classification, where the important visual patterns may be small, scattered, or difficult to identify from the full image alone. This is particularly valuable in medical image analysis, as the highlighted prediction-relevant regions can improve prediction transparency and provide clinicians with additional visual support during examination, screening, and case prioritization. Beyond medical applications, the same idea can also be applied to agricultural disease detection and other visual inspection tasks where localized abnormal patterns need to be identified. The framework can therefore serve as an assistive decision-support tool by providing lesion-focused visual evidence and classification support for expert assessment.

\section{Conclusion}\label{sec6}

In this study, we present a three-branch attention-driven architecture designed for lesion/spot-focused image classification. The architecture captures lesion-relevant cues while preserving global contextual information and adaptively selects the most informative representation for each sample. 
This leads to a stable and effective spot-based classification, particularly in scenarios where discriminative cues are small, scattered, and spatially dispersed.
We performed ablation studies to assess the choice of model components and hyperparameters. The proposed model was evaluated on four datasets and demonstrated consistent performance improvements over baseline models. In future studies, we plan to further improve the framework by exploring advanced preprocessing strategies, noise-aware attention mechanisms, and additional validation on larger independent datasets.

\section*{Declaration of competing interests}

The authors declare that they have no known competing financial interests or personal relationships that could have appeared to influence the work reported in this paper.

\section*{Funding sources}

This research did not receive any specific grant from funding agencies in the public, commercial, or not-for-profit sectors.

\section*{Data availability}

This study utilized one synthetically generated dataset and three publicly available datasets. The synthetic dataset was created solely for research purposes and is available from the corresponding author upon reasonable request. The publicly available datasets used in this study can be accessed through their respective original sources as cited in the references.

\section*{Declaration of generative AI and AI-assisted technologies in the manuscript preparation process}

During the preparation of this manuscript, the authors used generative AI tools solely for improving the clarity and readability of the language. The authors reviewed and edited the content as needed and take full responsibility for the content of the publication.

\bibliographystyle{elsarticle-num}
\bibliography{correct}

\end{document}